\documentclass{article}

\usepackage{arxiv}

\usepackage{url}

\usepackage{booktabs}
\usepackage[backend=biber,bibstyle=ieee]{biblatex}
\usepackage{acronym}

\usepackage{amssymb}

\usepackage{placeins}

\usepackage{graphicx}
\usepackage{svg}

\usepackage{kbordermatrix}

\usepackage{multicol}

\usepackage{mathtools} 
\usepackage{algorithm}
\usepackage{algpseudocode}

\usepackage{physics}
\usepackage{amsmath}
\usepackage{bm}
\renewcommand\vec[1]{\bm{#1}} % {\boldsymbol{\mathrm{#1}}}

\usepackage{subcaption}

\usepackage{nicefrac}

\usepackage{enumitem}

\newcommand\dvec[1]{\dot{\vec{#1}}}
\newcommand\hvec[1]{\hat{\vec{#1}}}
\newcommand\uvec[1]{\underline{\vec{#1}}}

\newcommand\eqblt{\mathrel{\overset{\makebox[0pt]{\mbox{\normalfont\tiny\sffamily BLT}}}{=}}}
\newcommand\eqdef{\mathrel{\overset{\makebox[0pt]{\mbox{\normalfont\tiny\sffamily $\Delta$}}}{=}}}

\usepackage{bm}
\renewcommand\vec[1]{\bm{#1}} % bold vectors

\newcommand{\githuburl}{\url{https://github.com/ThummeTo/HUDADE.jl}}

\newcommand{\sysa}{$\vec{s}_a$}
\newcommand{\sysb}{$\vec{s}_b$}
\newcommand{\sysz}{$\vec{s}_z$}

\newcommand\myfigure[3]{
    \begin{figure}[ht]
    \vskip 0.2in
        \begin{center}
            \centerline{\includegraphics[width=\columnwidth]{#1}}
            \caption{#2}
            \label{fig:#3}
        \end{center}
    \vskip -0.2in
    \end{figure}
}

\newcommand\myfigurewidth[4]{
    \begin{figure}[ht]
    \vskip 0.2in
        \begin{center}
            \centerline{\includegraphics[width=#4]{#1}}
            \caption{#2}
            \label{fig:#3}
        \end{center}
    \vskip -0.2in
    \end{figure}
}

\newcommand\myfiguremode[5]{
    \begin{figure}[#5]
    \vskip 0.2in
        \begin{center}
            \centerline{\includegraphics[width=#4]{#1}}
            \caption{#2}
            \label{fig:#3}
        \end{center}
    \vskip -0.2in
    \end{figure}
}

\usepackage{tabularx}

\usepackage{draftwatermark}
\SetWatermarkText{PREPRINT}
\SetWatermarkScale{1}
\definecolor{wm}{gray}{0.95}
\SetWatermarkColor{wm}

\title{Learnable \& Interpretable Model Combination\\ in Dynamical Systems Modeling}

\author{%
	Tobias Thummerer \\
	Chair of Mechatronics\\
    University of Augsburg\\
    86159 Augsburg, Germany\\
	\texttt{tobias.thummerer@uni-a.de}  
	\And 
	Lars Mikelsons \\
	Chair of Mechatronics\\
    University of Augsburg\\
    86159 Augsburg, Germany\\
	\texttt{lars.mikelsons@uni-a.de}  
}
\date{}

\renewcommand{\headeright}{Preprint}
\renewcommand{\undertitle}{Preprint}
\renewcommand{\shorttitle}{~}

\begin{document}

% don't print acronyms long:
\acused{HUDA}
\acused{GPU}

\maketitle

\begin{abstract}
  During modeling of dynamical systems, often two or more \emph{model architectures} are combined to obtain a more powerful or efficient model regarding a specific application area. This covers the combination of multiple machine learning architectures, as well as \emph{hybrid models}, i.e., the combination of physical simulation models and machine learning.
    In this work, we briefly discuss which types of model are usually combined in dynamical systems modeling and propose a class of models that is capable of expressing mixed algebraic, discrete, and differential equation-based models. Further, we examine different established, as well as new ways of combining these models from the point of view of system theory and highlight two challenges - algebraic loops and local event functions in discontinuous models - that require a special approach. Finally, we propose a new wildcard architecture that is capable of describing arbitrary combinations of models in an easy-to-interpret fashion that can be learned as part of a gradient-based optimization procedure. In a final experiment, different combination architectures between two models are learned, interpreted, and compared using the methodology and software implementation provided.
\end{abstract}

\twocolumn

\section{Introduction}
One of the core concepts in science, and something that happens intuitively in everyday modeling of dynamic systems, is the \emph{combination} of models or methods. With the goal of further optimizing regression and classification models of dynamic systems, models are combined in a way that merges their positive characteristics. This comes in a variety of different flavors: For example, in \cite{Gu:2021}, recurrent, convolutional, and linear continuous-time differential equations are combined to improve speech classification. Furthermore, \cite{Rubanova:2019} introduces the combination of a (neural) \ac{ODE}, a \ac{RNN} and an additional \ac{FFNN}, called \ac{ODE}-\ac{RNN}, that improves the capability of pure \ac{RNN} of handling irregular sampled data. In the domain of \ac{SciML}, machine learning structures are paired with powerful algorithms from scientific computing. One example for this approach are neural \acp{ODE}, which are neural networks that are solved by numerical integration, as introduced in \cite{Chen:2018}. Training neural event functions to capture discontinuous dynamics is also possible, as \cite{Chen:2021} shows. In \emph{hybrid modeling}, existing simulation models can be extended by an additional machine learning part (or the other way around), often to learn for unknown and/or not modeled physical effects, such as, for example, the change in arterial cross-section by pressure in the human cardiovascular system, as in \cite{Thummerer:2021}. The list could go on and on because, apart from a few basic models, almost all model types in \ac{ML} can be expressed as a combination of two or more existing model types. Combining models is sometimes trivial and is performed intuitively without explicitly thinking about it. However, for (mathematically) more complex model combinations, new problems arise, as discussed in detail in Sec. \ref{sec:sota}. This raises the questions of \emph{how} a model is defined and what a proper \emph{model combination} looks like.

The meaning of the term \emph{model} differs between the mentioned application domains: Deep \acp{FFNN} for example, describe a nonlinear mapping from the input of the first layer $\vec{u}$ to the output of the last layer $\vec{y}$. \acp{RNN} on the other hand, also depend on the internal \emph{hidden state} of the architecture. Common simulation models of dynamic systems (such as from physics or economics) are often available as \emph{time-step-models}, which predict the next state $\vec{x}(t+\Delta t)$ based on the current state $\vec{x}(t)$ by doing one step $\Delta t$ in time. It is also common to use \acp{ODE}, that map from the current state $\vec{x}(t)$ to the state derivative $\dvec{x}(t)$, without doing a step in time. These models need to be paired (or \emph{combined}) with a numerical solver for differential equations for inference. In summary, we need to consider a wide variety of models (or model types) to investigate the \emph{combinability of models} in the large field of dynamical systems modeling.

In the case of combining pure machine learning models, considering how much is known about the interface between two models, we often have a good intuition of \emph{how} the architectures should interact in the later application. This results in a straightforward implementation of the connection. Unlikely, if we combine simulation models and \emph{machine learning}, a very good understanding of the interaction between both models is required. The more complex simulation models become, the greater the uncertainty about what the perfect interface for extending it with \ac{ML} is. Especially in \emph{such} cases, a way of defining a generic interface between models that can be learned is desirable.

From the implementation point of view, \emph{combination of models} is performed and validated for the specific application area or the specific use case of the model. This often results in the development and publication of a dedicated algorithm for inference (or solving) and training the \ac{CM}, as well as a new software library that implements the proposed methods. Providing (unnecessary) new implementations is error-prone and often results in suboptimal computational performance or lacking hardware support (like e.g. for \acp{GPU}). A common super class for \acp{CM} will allow providing not only a common methodology to deal with these models, like e.g. advanced training methods, but also a versatile software foundation.

After providing a short state-of-the-art overview (Sec. \ref{sec:sota}) and summarizing the problem statement, the methodological contribution is separated into two parts: In Section \ref{sec:model}, we briefly introduce a common model class and interface. In Section \ref{sec:paradigm}, we discuss topologies to combine these models (statically or learnable), issues that can result from that combination, and solutions for these issues.

\section{State of the art}\label{sec:sota}
The topic of model combination is probably best discussed in the context of \emph{hybrid modeling}, also known as \emph{gray-box modeling}. This field of research deals with the combination of physical simulation models, further referred to as \acp{FPM} and \acp{MLM}. Publications like \cite{VonRueden:2020} discuss the combinability of such models in a broad context, concerning possibilities on where and how different models can interact along the application. Numerous other publications such as \cite{Thompson:1994,Sohlberg:2008,Rudolph:2024} (to name only a tiny subset) go one step further and examine how simulation models and \ac{ML} can actually be combined \emph{structurally}, leading to the two most common architecture patterns of the \emph{serial} (s. Fig. \ref{fig:serial}) and \emph{parallel} (s. Fig. \ref{fig:parallel}) topology. Finally, there are dedicated solutions for specific application domains such as thermodynamic fluid mixtures \cite{Jirasek:2023} or ocean vessels \cite{Leifsson:2008}, again to name only a few.

\myfigurewidth{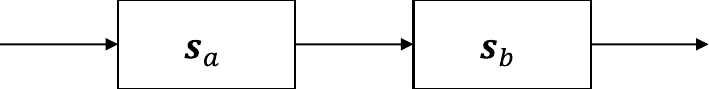}{The serial topology, as proposed in \textcite{Thompson:1994} or further discussed, e.g. in \textcite{Rudolph:2024}. The submodel \sysa{} processes values, before passing them further to subsystem \sysb{}, that computes the actual results.}{serial}{6cm}

\myfigurewidth{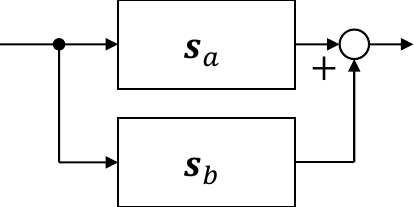}{The parallel topology, as proposed in \textcite{Thompson:1994} or further discussed, e.g. in \textcite{Rudolph:2024}. The submodel \sysb{} learns for the residuals of subsystem \sysa{} and adds them to the common result.}{parallel}{3.5cm}

\paragraph{Open Challenge}
It is quite easy to show that the proposed structures are often not feasible in more complex, real applications. This can be traced back to algebraic loops (s. Fig. \ref{fig:serial_fail1}) and non-trivial event handling (s. Fig. \ref{fig:serial_fail2}) within the \ac{CM}. This implies that no model inference (and no training) is possible without further methodological adaptations.

\myfigurewidth{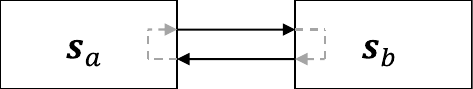}{For submodels \sysa{} and \sysb{} with an algebraic relation between inputs and outputs (for example, if both are \acp{FFNN}), algebraic loops can easily be constructed by connecting both models cyclically. Without handling this algebraic loop, the resulting \ac{CM} can't be evaluated (inference and backpropagation) correctly.}{serial_fail1}{4cm}
	
\myfigurewidth{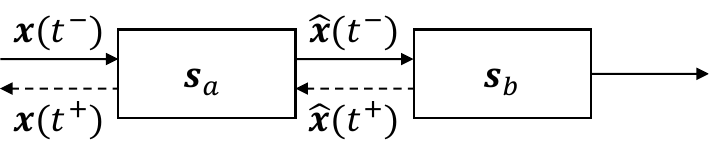}{During regular simulation (without events, solid arrows), submodel \sysa{} maps the state $\vec{x}(t^-)$ to $\hvec{x}(t^-)$, submodel \sysb{} computes further quantities (for example the state derivative) based on this. In case of an event in the submodel \sysb{} (dashed arrows), the local state within \sysb{} $\hvec{x}(t^-)$ is updated to $\hvec{x}(t^+)$. This new local state must be propagated backwards through submodel \sysa{} to obtain a new global state $\vec{x}(t^+)$ to proceed with the numerical integration.}{serial_fail2}{6cm}

The reason behind this is that the state of the art captures the \emph{theoretical possibilities} of the model combination in \ac{ML}, but lacks statements on the \emph{mathematical consequences} of it, such as the solvability of the \ac{CM}. The topic of model combination must be investigated on the level of systems of equations, considering the different types of equations that can occur during the modeling of dynamical systems in \ac{ML}. It is necessary to not only discuss \emph{how} models could be combined, but how they can be combined \emph{meaningfully} - so without construction of additional algebraic loops and proper handling of discontinuities. In the introduced context, this step has not yet been taken to the knowledge of the authors. In addition, the combination of models in a (almost) fully learnable way instead of statically connecting them, as well as the mathematical consequences of these \emph{learnable connections} have also not been studied yet.

\section{Definition: \acs{HUDA}-\acsp{ODE}}\label{sec:model}
As stated, a common interface for dynamical system models is needed in order to discuss the combinability of such. In the following, we briefly introduce the class of \acl{HUDA} \aclp{ODE} (\acs{HUDA}-\acsp{ODE}), that is able to express the model types considered from the introduction. However, this class should not be misunderstood as a novel mathematical structure, but as a system definition to investigate and solve the problem considered. As a direct consequence, \acs{HUDA}-\acsp{ODE} can be solved by any single- or multistep \ac{ODE} solver with support for event identification and handling. 

We start with the foundation of an \ac{ODE}, that is capable of describing dynamic processes:
\begin{equation}\label{equ:f_old}
	\dvec{x}(t) = \vec{f}(\vec{x}(t), \vec{u}(t), \vec{p}, t)
    .
\end{equation} 
The \ac{ODE} depends on the system state $\vec{x}$, parameter vector $\vec{p}$, input $\vec{u}$ and time $t$. Technically, inputs can be implemented as parameters (if constant), by augmentation of the continuous states (if the input derivatives are known), implemented as discrete states that are changed by time-events (sampled inputs) or the input equation (if known) can be added to the system of equations, in a way that the system boundary is extended (and no explicitly defined input is needed). Because $\vec{f}$ may also include \acp{UA}, one might also refer to it as \ac{UODE} as introduced in \cite{Rackauckas:2021}. However, we don't use the \ac{UA} as explicit function argument here and only state that the right-hand side may include \acp{UA} and the parameter vector $\vec{p}$ may include physical as well as \ac{ML} parameters. To further allow for modeling discontinuous (or hybrid) systems, the \ac{UODE} can be extended to the \emph{hybrid} \ac{UODE} by providing an \emph{event condition} function $\vec{c}$, indicating events by zero crossings:
\begin{equation}
	\vec{z}(t) = \vec{c}(\vec{x}(t), \vec{u}(t), \vec{p}, t) 
    .
\end{equation}
In the following, we express the time\footnote{This is a simplified introduction of the concept of \emph{super-dense time}, see \cite{FMI:2020, FMI:2023} for more information.} after the event happened as $t^+$ and the time right before the event as $t^-$. In analogy, time dependent variables $\phi$ (like e.g. the state) are abbreviated $\phi^+$ and $\phi^-$ for $\phi(t^+)$ and $\phi(t^-)$. The \emph{event affect} function $\vec{a}$ computes the state after the event based on the state before the event and is defined as
\begin{equation}
    \vec{x}(t^+) \eqdef
    \vec{x}^+ = 
    \vec{a}(\vec{x}^-, \vec{u}^-, \vec{p}, t^-, \vec{q})
    ,
\end{equation}
where $\vec{q}$ is a multi-hot vector identifying zero crossings ($1$) and no zero crossings ($0$) of the event indicators. To further allow for purely algebraic mappings (for example in \acp{FFNN}), it is common to augment (U)\acp{ODE} by an algebraic \emph{output} equation, given as
\begin{equation}
	\vec{y}(t) = \vec{g}(\vec{x}(t), \vec{u}(t), \vec{p}, t)
    .
\end{equation}
Further, discrete subsystems, that only change for discrete points in time (like \acp{RNN}), need to be considered. The state vector $\vec{x}$ can be separated into the \emph{continuous} state $\vec{x}_c$, that changes continuously over time (by the given derivative in $\vec{f}_c$), and the \emph{discrete} state $\vec{x}_d$, that changes only at event instances (inside the function $\vec{a}$). The original equation $\vec{f}$ (s. Eq. \ref{equ:f_old}) can therefore be separated into two equations and reformulated:
\begin{equation}
	\dvec{x}(t) = 
	\begin{bmatrix}
		\dvec{x}_c(t) \\ 
		\dvec{x}_d(t)
	\end{bmatrix} = 
	\begin{bmatrix}
		\vec{f}_c(\vec{x}_c(t), \vec{x}_d(t), \vec{u}(t), \vec{p}, t) \\
		\vec{0}
	\end{bmatrix}
    .
\end{equation}
The complete \acs{HUDA}-\ac{ODE} reads:
\begin{equation}
	\begin{bmatrix}
		\dvec{x}_c(t) \\ 
		\dvec{x}_d(t) \\ 
		\vec{y}(t) \\
		\vec{z}(t) \\
		\vec{x}(t^+)
	\end{bmatrix} = 
	\begin{bmatrix}
		\vec{f}_c(\vec{x}_c(t), \vec{x}_d(t), \vec{u}(t), \vec{p}, t) \\
		\vec{0} \\
		\vec{g}(\vec{x}_c(t), \vec{x}_d(t), \vec{u}(t), \vec{p}, t)\\
		\vec{c}(\vec{x}_c(t), \vec{x}_d(t), \vec{u}(t), \vec{p}, t)\\
		\vec{a}(\vec{x}_c(t^-), \vec{x}_d(t^-), \vec{u}(t^-), \vec{p}, t^-)
	\end{bmatrix}
    .
\end{equation}
To summarize, the \ac{HUDA}-\ac{ODE} is fully contained within the class of \acp{UODE}. Its purpose is to define a model structure and interface for model combination, that can be applied to a significant subset of dynamical system models from \ac{ML} and physical systems modeling. It consists of five subsystems of equations with associated meaning. All parts of the \ac{HUDA}-\ac{ODE} might be derived from first principles or be \ac{ML} models (or a mixture of both). 

This definition of the model was not chosen arbitrarily, but was strongly motivated by the \ac{FMI} \cite{FMI:2020, FMI:2023}. Whereas \ac{FMI} is a software standard, the mathematical aspects within are very closely related. As a direct consequence, Functional Mock-Up Units, thus models that implement the \ac{FMI}, fall into this mathematical class. This further makes the presented research directly applicable to industrial simulation models via \ac{FMI}.

\section{Combining \acs{HUDA}-\acsp{ODE}}\label{sec:paradigm}
Before discussing the combinability of \ac{HUDA}-\acp{ODE}, we observe some basic principles that lead to a simplified discussion:
\begin{itemize}
    \item  Except for the \ac{ODE} $\vec{f}_c$, all systems of equations are pure algebraic equations. If $\dvec{x}_c$ is interpreted as unknown (instead of a derivative that is numerically integrated), even the equation $\vec{f}_c$ can be handled as an algebraic equation. This insight allows us to discuss only algebraic systems in the following. 
	\item It is sufficient that we discuss the combination of exactly two algebraic systems and show that this leads to an algebraic system again. In this way, we can easily extend the concept to the arbitrary case of recursively combining multiple algebraic systems (or \ac{HUDA}-\acp{ODE}).
	\item It is sufficient to consider only a single vector input $\vec{\upsilon}$ and a single vector output $\vec{\gamma}$ for the models, because multiple inputs/outputs can be concatenated into a single vector. Even matrix or tensor inputs can be flattened and expressed as a vector. The same applies if we investigate multiple systems of equations: We can concatenate (augment) them into a single one.
\end{itemize}
As a consequence, for a simplified deduction of the combinability of \ac{HUDA}-\acp{ODE}, we can express them as a single system of algebraic equations with single vectorial input and single vectorial output.

\subsection{Topology: Generic}
The combination of two algebraic subsystems $\vec{s}_a$ and $\vec{s}_b$ - note, that this could be any of $\vec{f}_c$, $\vec{g}$, $\vec{c}$ or $\vec{a}$ of a \ac{HUDA}-\ac{ODE}, a subset or even all of them - into the new algebraic equation system $\vec{s}_z$ can be mathematically expressed as the concatenation of the two systems of equations, as visualized in Fig. \ref{fig:general}. However, new unknowns are introduced for the inputs of the submodels $\vec{\upsilon}_a$ and $\vec{\upsilon}_b$ and the output of the combined model $\vec{\gamma}_z$.

\myfigurewidth{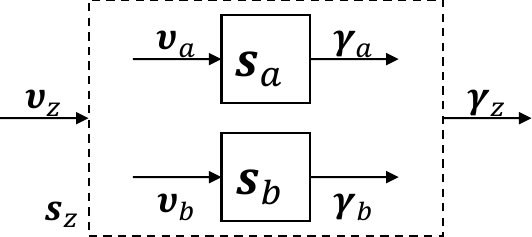}{Two algebraic subsystems $\vec{s}_a$ and $\vec{s}_b$ can be combined to a single system $\vec{s}_z$. Because new unknowns $\vec{\upsilon}_a$, $\vec{\upsilon}_b$ and $\vec{\gamma}_z$ are introduced, the system is not well-posed. This can be seen in the missing connections between the input of the combined model ($\vec{\upsilon}_z$) and its output ($\vec{\gamma}_z$) and the inputs of the submodels ($\vec{\upsilon}_a$ and $\vec{\upsilon}_b$) and the outputs ($\vec{\gamma}_a$ and $\vec{\gamma}_b$).}{general}{7cm}

To gain a solvable system, we augment the system of equations (consisting of submodels \sysa{} and \sysb{}) by three linear equations $\vec{c}_a$, $\vec{c}_b$ and $\vec{c}_z$, connecting the newly introduced unknowns and summarize:
\begin{align}
    \vec{\gamma}_a &= \vec{s}_a(\vec{\upsilon}_a) \\ %\qquad\qquad \text{\textcolor{gray}{(submodel $a$)}}\\
 	\vec{\gamma}_b &= \vec{s}_b(\vec{\upsilon}_b) \\ %\qquad\qquad\> \text{\textcolor{gray}{(submodel $b$)}}\\
	\vec{\upsilon}_a &= \vec{c}_a(\vec{\gamma}_a, \vec{\gamma}_b, \vec{\upsilon}_z) \notag\\
    &= \vec{W}_{aa} \vec{\gamma}_a + \vec{W}_{ab} \vec{\gamma}_b + \vec{W}_{az} \vec{\upsilon}_z + \vec{b}_a \label{equ:ca} \\
	\vec{\upsilon}_b &= \vec{c}_b(\vec{\gamma}_a, \vec{\gamma}_b, \vec{\upsilon}_z) \notag\\
    &= \vec{W}_{ba} \vec{\gamma}_a + \vec{W}_{bb} \vec{\gamma}_b + \vec{W}_{bz} \vec{\upsilon}_z + \vec{b}_b \label{equ:cb} \\
	\vec{\gamma}_z &= \vec{c}_z(\vec{\gamma}_a, \vec{\gamma}_b, \vec{\upsilon}_z) \notag\\
    &= \vec{W}_{za} \vec{\gamma}_a + \vec{W}_{zb} \vec{\gamma}_b + \vec{W}_{zz} \vec{\upsilon}_z + \vec{b}_z \label{equ:cz} 
    ,
\end{align}
with connection matrix $\vec{W} \in \mathbb{R}^{n \times m}$ and connection bias $\vec{b} \in \mathbb{R}^{n}$
\begin{equation}
	\vec{W} = 
	\kbordermatrix{
		~         & \vec{\gamma}_a    &\vec{\gamma}_b     & \vec{\upsilon}_{z}  \\
		\vec{\upsilon}_a & \vec{W}_{aa} & \vec{W}_{ab} & \vec{W}_{az} \\
		\vec{\upsilon}_b & \vec{W}_{ba} & \vec{W}_{bb} & \vec{W}_{bz} \\
		\vec{\gamma}_z & \vec{W}_{za} & \vec{W}_{zb} & \vec{W}_{zz} \\
	}
    \quad 
    \vec{b} = 	
	\kbordermatrix{
		~         & ~    \\
		\vec{\upsilon}_a & \vec{b}_{a} \\
		\vec{\upsilon}_b & \vec{b}_{b} \\
		\vec{\gamma}_z & \vec{b}_{z} \\
	}
    ,
\end{equation}
where $n = |\vec{\upsilon}_a|+|\vec{\upsilon}_b|+|\vec{\gamma}_z|$ and $m = |\vec{\gamma}_a|+|\vec{\gamma}_b|+|\vec{\upsilon}_z|$. The row markings correspond to the equations \ref{equ:ca}-\ref{equ:cz} and the column markings correspond to the function inputs. In this way, we can intuitively interpret which part of the matrix $\vec{W}$ corresponds to which variable, for example $\vec{W}_{zb}$ connects to "z" (\sysz{} output $\vec{\gamma}_z$), coming from "b" (\sysb{} output $\vec{\gamma}_b$).

These new linear equations are fully differentiable with respect to their parameters $\vec{W}$ and $\vec{b}$, which is a prerequisite for gradient-based optimization in differentiable programming, and can be used in multiple ways:
\begin{enumerate}[label=(\alph*)]
	\item The entries of the matrix $\vec{W}$ can be set to identify a connection ($W_{ij}=1$) or no connection ($W_{ij}=0$) between two variables, while keeping the bias $\vec{b}$ at zeros. This is the common understanding of a connection matrix. By providing a non-permutation matrix for $\vec{W}$, we can further \emph{mix} multiple signals. Signals can be amplified ($W_{ij}>0$), attenuated ($W_{ij}<1$), inverted ($W_{ij}<0$) or feature offsets ($b_{i} \neq 0$). 
	\item The entries of $\vec{W}$ and $\vec{b}$ can be initialized with predetermined values. This way, the equations can be used as linear pre- and post-processing layers as proposed in \cite{Thummerer:2022}, to scale and shift signals between the combined models to an appropriate value range (e.g. a normal distribution for \acp{ANN}).
	\item The parameters can be initialized by applying (a) or (b) and be further optimized together with other parameters. This is done by augmenting the parameter vector of the resulting \ac{HUDA}-\ac{ODE}. The resulting combined model is capable of describing arbitrary connections between two given algebraic subsystems - and \ac{HUDA}-\acp{ODE} respectively - expressed by its parameterization.
\end{enumerate}
Note, that besides \emph{addition}, these connection equations are able to express \emph{concatenation} of signals as well, dependent on the chosen dimension\footnote{For example, if it states that $|\vec{\gamma}_z| = |\vec{\gamma}_a| + |\vec{\gamma}_b|$, the connection matrices $\vec{W}_{za}$ and $\vec{W}_{zb}$ can be populated so that the upper part of $\vec{\gamma}_z$ contains the entries of $\vec{\gamma}_a$ and the lower part the entries of $\vec{\gamma}_b$ respectively.} of the combined model input $\vec{\upsilon}_z$ and output $\vec{\gamma}_z$. Furthermore, if only some parameters within $\vec{W}_{za}$ and $\vec{W}_{zb}$ are trainable, we can implement \emph{gates} to optimize (or control) signal gains within the architecture \cite{Thummerer:2022}. 

By introducing the connection equations, the combined system is balanced and the corresponding incidence matrix $\vec{D}$ reads:
\begin{equation}\label{eq:D}
	\vec{D} = 
	\kbordermatrix{
		~             & \vec{\gamma}_a    & \vec{\gamma}_b    & \vec{\upsilon}_a    & \vec{\upsilon}_b    & \vec{\gamma}_z\\
		\vec{s}_a     & \uvec{I}     & \vec{0}      & \vec{D}_{a}  & \vec{0}      & \vec{0}  \\
		\vec{s}_b 	  & \vec{0}      & \uvec{I}     & \vec{0}      & \vec{D}_{b}  & \vec{0}  \\
		\vec{c}_{a}   & \vec{D}_{aa} & \vec{D}_{ab} & \uvec{I}     & \vec{0}      & \vec{0}  \\
		\vec{c}_{b}   & \vec{D}_{ba} & \vec{D}_{bb} & \vec{0}      & \uvec{I}     & \vec{0}  \\
		\vec{c}_z     & \vec{D}_{za} & \vec{D}_{zb} & \vec{0}      & \vec{0}      & \uvec{I} \\
	}
    .
\end{equation}
The incidence matrix $\vec{D}$ holds the unknown\footnote{In terms of variables that we solve for.} variables in columns and the equations in rows (compare marks). The diagonal indicates which equation is solved for which variable. It can be interpreted as a simplified Jacobian of the equations (rows) regarding the variables (columns), where zero entries are kept zero, but nonzero entries are simplified with ones. Such matrices are used to investigate the \emph{dependencies} within a system of equations, allowing us to make statements regarding sparsity, singularity, and algebraic loops. The incidence matrix for a connection submatrix $\vec{W}_{ij}$ is indicated by $\vec{D}_{ij}$, where $i,j \in \{a,b,z\}$. Note that only the incidence submatrices for a part of the connection matrix $\vec{W}$ are included, because the last column of $\vec{W}$ refers to the \emph{known} variable $\vec{\upsilon}_{z}$ and the columns in the incidence matrix only hold \emph{unknowns}. In addition, the incidence matrices of the subsystems to be combined $\vec{D}_a$ and $\vec{D}_b$ are included. Without further arrangements and for nontrivial incidence matrices $\vec{D}_a \neq \vec{0}$ and $\vec{D}_b \neq \vec{0}$, the system behind $\vec{D}$ contains algebraic loops. This can easily be proven by trying to perform a \ac{BLT} transformation on Eq. \ref{eq:D} - which is not possible, because all rows contain at least two nonzero entries. Algebraic loops do not prevent solving of the system, but require appropriate handling by identification (e.g., Tarjan's algorithm \cite{Tarjan:1972}) and solving (e.g., Newton's method). In case of optimizing the parameters of $\vec{W}$, identification needs to be repeated after every optimization step, and solving within every model inference, which is quite costly in terms of computational performance. Here, a topology free of algebraic loops (caused by combination) might be desirable and is therefore derived as follows.

\subsection{Topology: Parallel-Sequential-DFT (PSD)}
To prevent algebraic loops, it is clear that for arbitrary $\vec{D}_a$ and $\vec{D}_b$, connections of the equation blocks $\vec{s}_a$ and $\vec{s}_b$ with themselves must be prohibited, which results in $\vec{W}_{aa} = \vec{W}_{bb} = \vec{0}$. In addition, we need to avoid cyclical connections between both submodels, which we can achieve by forcing one submodel to be evaluated \emph{before} the other. This can be achieved by dictating $\vec{W}_{ab} = \vec{0} \lor \vec{W}_{ba} = \vec{0}$. Finally, we notice that the \ac{DFT} between \ac{CM} input and output of \ac{CM} cannot create algebraic loops and therefore the corresponding part of the connection matrix $\vec{W}_{zz}$ is kept for the combined model. To summarize, exactly two meaningful cases remain, further referred to as case $a$ and case $b$, where a system free of algebraic loops\footnote{Free of algebraic loops caused by the combination scheme, if there are already algebraic loops inside of submodel \sysa{} or \sysb{}, they remain of course.} can be constructed, so a \ac{BLT} transformation is possible:
\begin{enumerate}[label=(\alph*)]
	\item $\vec{W}_{aa} = \vec{W}_{bb} = \vec{W}_{ab} = \vec{0}$ (see the following deduction) or
	\item $\vec{W}_{aa} = \vec{W}_{bb} = \vec{W}_{ba} = \vec{0}$ (see appendix section \ref{sec:appendix_caseb}).
\end{enumerate}
Both cases of this topology can be visualized as in Figs. \ref{fig:loop_free_a} and \ref{fig:loop_free_b} (Appendix). Because this topology has parallel (P), sequential (S) and \ac{DFT} (D) characteristics, it can be referred to as topology \emph{PSDa} (for case a) and \emph{PSDb} (for case b).

\myfigurewidth{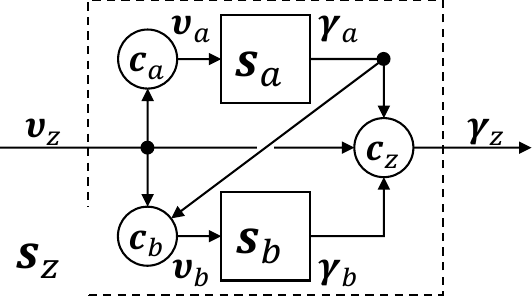}{\emph{PSDa}: $\vec{s}_a$ is evaluated before $\vec{s}_b$, resulting in $\vec{W}_{ab} = \vec{0}$. The connection equations $\vec{c}_a$, $\vec{c}_b$ and $\vec{c}_z$ are visualized. In between, each arrow corresponds to a part of the connection matrix $\vec{W}$. For example, the diagonal arrow (center) aims on "b" (submodel \sysb{}), coming from "a" (submodel \sysa{}) and therefore corresponds to the submatrix $\vec{W}_{ba}$.}{loop_free_a}{7cm}

For the \emph{PSDa} topology, the system incidence matrix, that is now more sparse compared to the generic topology, can be transformed to \ac{BLT} form, and therefore corresponds to a solvable system:
\begin{align}\label{equ:D_PSD}
	\vec{D}_{PSDa} 
    &\eqblt
	\kbordermatrix{
		~             & \vec{\upsilon}_a    & \vec{\gamma}_a    & \vec{\upsilon}_b    & \vec{\gamma}_b    & \vec{\gamma}_z\\
		\vec{c}_{a}   & \uvec{I}     & \vec{0}      & \vec{0}      & \vec{0}      & \vec{0}  \\
		\vec{s}_a     & \vec{D}_{a}  & \uvec{I}     & \vec{0}      & \vec{0}      & \vec{0}  \\
		\vec{c}_{b}   & \vec{0}      & \vec{D}_{ba} & \uvec{I}     & \vec{0}      & \vec{0}  \\
		\vec{s}_b 	  & \vec{0}      & \vec{0}      & \vec{D}_{b}  & \uvec{I}     & \vec{0}  \\
		\vec{c}_z     & \vec{0}      & \vec{D}_{za} & \vec{0}      & \vec{D}_{zb} & \uvec{I} \\
	} 
    .
\end{align}
The proposed PSD topology allows for the combination of two algebraic subsystems (with unknown internal dependencies $\vec{D}_a$ and $\vec{D}_b$) in the most generic way, such that the resulting system is free of algebraic loops by design. The only decision that must be made in advance is which subsystem is evaluated first, resulting in the two cases (a) and (b). One subsystem is evaluated first, then the second subsystem is evaluated based on the intermediate results of the first, as well as on the \ac{CM} input $\vec{\upsilon}_z$. The \ac{CM} output depends on results from both submodels and the \ac{DFT} of the \ac{CM} input $\vec{\upsilon}_z$. 

Of course, simpler topologies, including the known concepts of serial and parallel structure, can be derived from the \emph{PSD} topology by setting further parts of the connection matrix $\vec{W}$ to zeros. Note that we only need to distinguish between cases (a) and (b) for topologies including the serial pattern. Otherwise, both submodels can be evaluated in arbitrary order or at once. In table \ref{tab:topologies} (Appendix) and the following, topologies are distinguished by naming only the attributes (or connections) that are applied. For example, \emph{PSa} identifies the topology that uses the P(arallel) and S(equential) layout for case (a) (\sysa{} is evaluated before \sysb{}), but not the D(FT) connections. The topology with only zero connections would refer to an empty identifier but is not of practical relevance and not investigated further.

\subsection{Local state affect function}
Event handling of combined systems is challenging, for two reasons:\begin{enumerate}
    \item During an event instance, the \textbf{causalization changes}. States are usually solved by numerical integration and thus are input to the system of equations and output of the \ac{ODE} solver. However, in case of an event, the state is determined by the event affect function of the system, and is therefore output of the system and input to the \ac{ODE} solver.
    \item The submodels feature a \textbf{local definition of state}. If the local state of a submodel changes at an event instance, this local state needs to be propagated through the entire combined model to finally obtain a new \emph{global} state that can be passed to the \ac{ODE} solver. 
\end{enumerate}
% \emph{local state} of the subsystem, this means the state definition is only valid for the submodel and not the \emph{global state} of the \ac{CM}. This further results in a \emph{local} event affect function: The new state for after the event instance corresponds to the \emph{local state} definition of the subsystem that triggered the event. 
% This \emph{local state} does not match the \emph{global state} of the combined system and can even have another dimension, because of the introduced linear connection equations. 
For a better understanding, we investigate this in more detail. It is important to keep in mind that both submodels share a single \ac{ODE} solver for the \ac{CM}. During regular simulation, the global state of the system is solved by the \ac{ODE} solver and is input to the \ac{CM}. Within the \ac{CM}, the local states of the submodels are determined by evaluating the connections and other submodels. At the time of event handling, the causalization of the system of equations changes: Instead of \emph{providing} a global state, the \ac{ODE} solver requires a new state after the event to be given. However, the submodel that triggered the event only provides a local state that is valid within the submodel only. A global state must be determined that matches the new local state for the event subsystem, and thus we derive a dedicated event handling routine for this in the following. 

In the following, we investigate the case of an event in the subsystem \sysb{} for the \emph{PSDa} topology. Events in subsystem \sysa{} for \emph{PSDa} are handled more easily, because the new local state only needs to be propagated backwards through a single connect equation. This is described in Sec. \ref{sec:appendix_eventa} (appendix). To make it easier to follow, but without limiting the validity, we assume in the following that only states are considered as inputs and state derivatives as outputs, thus $\vec{\upsilon}_{\phi} = \vec{x}_{\phi}$ and $\vec{\gamma}_{\phi} = \dvec{x}_{\phi}$ for $\phi \in \{a,b,z\}$. For \emph{PSDa}, the incidence matrix of the system at event in \sysb{} states:
\begin{equation}\label{equ:D_event_b}
	\vec{D}_{PSDa}^{(b)} \eqblt
	\kbordermatrix{
		~             & \vec{\gamma}_b    & \vec{\gamma}_a    & \vec{\upsilon}_a    & \vec{\upsilon}_z    & \vec{\gamma}_z\\
		\vec{s}_b 	  & \uvec{I}     & \vec{0}      & \vec{0}      & \vec{0}      & \vec{0}  \\
		\vec{c}_{b}   & \vec{0}      & \uvec{D}_{ba}& \vec{0}      & \vec{D}_{bz} & \vec{0}  \\
		\vec{s}_a     & \vec{0}      & \vec{I}      & \uvec{D}_{a} & \vec{0}      & \vec{0}  \\
		\vec{c}_{a}   & \vec{0}      & \vec{0}      & \vec{I}      & \uvec{D}_{az}& \vec{0}  \\
		\vec{c}_{z}   & \vec{D}_{zb} & \vec{D}_{za} & \vec{0}      & \vec{D}_{zz} & \uvec{I} \\
	} 
    .
\end{equation}
Because this system contains an algebraic loop for any of the cases $\vec{D}_{bz} \neq \vec{0}$ and/or $\vec{D}_{ba}$, $\vec{D}_{a}$, $\vec{D}_{az}$ are not in \ac{BLT} shape, the solution must be obtained by solving a nonlinear system of equations. The corresponding residual $\vec{r}^{(b)}$ is derived by inserting equations $\vec{c}_a$, $\vec{s}_a$, and $\vec{c}_b$, into each other. This order is obtained by the \ac{BLT} transform in equation \ref{equ:D_PSD}. The residual states:
\begin{equation} % c_b
	\vec{r}^{(b)}(\vec{\upsilon}_z) = 
	\underbrace{\vec{W}_{ba}\vec{s}_a(\vec{W}_{az} \vec{\upsilon}_z)}_\text{via subsystem \sysa{}} + 
	\underbrace{\vec{W}_{bz} \vec{\upsilon}_z}_\text{via CM input} - 
	\vec{\upsilon}_{b}
    .
\end{equation}
Depending on the involved mathematical operations between global and local state, the solution might not be unique. Alternatively, this can be solved by optimization if we take the absolute value of the residual as minimization goal. During solving of the nonlinear system, it is important to change only the sensitive entries of $\vec{\upsilon}_z$, that is, the entries with nonzero gradient $\nicefrac{d \vec{r}^{(b)}}{d \vec{\upsilon}_z}$. Otherwise, independent parts of the global state vector might be unintentionally changed. 
For different edge cases, for example, invertible submodels and connection matrices, analytical solutions can be derived for the global state $\vec{\upsilon}_z$, but are omitted for reasons of space.
The cases for the topology \emph{PSDb} can be derived in a straightforward way from the cases presented for \emph{PSDa}.

\section{Experiment}
 The contributions of this paper are shown with the help of two experiments, we added a remark regarding the choice of experiment in the Appendix \ref{sec:remark}. First, we deploy the proposed architectures and train them for different \textbf{connection patterns} inside a \ac{CM}. Second, we show that besides the \ac{CM} in the previous example, other \acp{MLM} can be trained using the very same training loop, showing the versatility and usefulness of the \textbf{common interface} that is provided by the \ac{HUDA}-\ac{ODE}, see Sec. \ref{sec:common-interface} (Appendix). 
The corresponding software library, implemented in the Julia programming language \cite{Bezanson:2017}, as well as example scripts, can be found on GitHub\footnote{\githuburl{}}. 

\subsection{Training}
In the following experiment, we create a \ac{CM} by combining a \ac{FPM} and a \ac{MLM} by applying different of the proposed patterns. The six investigated patterns are generated by permutation of the attributes S(equential), P(arallel) and D(FT), excluding the cases without any connections and the topology $D$, which is by design not capable of expressing the bouncing ball\footnote{Note, that the topology $D$ lineraly connects the \ac{CM} inputs (so the state) with the outputs (state derivative).}. The \ac{FPM}, in form of an \ac{ODE}-based simulation model, represents the well-known bouncing ball in two dimensions ($x$ and $y$) with collision surfaces at $x=\{-1,1\}$ and $y=\{-1,1\}$, but without air friction. The \ac{MLM} is a simple \ac{FFNN} and its parameters are optimized together with the parameters of the connections. 

In addition, ground-truth data from a more detailed simulation of the bouncing ball, including a simple quadratic air friction model, is available and used for training and testing. For better comparability of the patterns, training is carried out with very little data (approx. $8$ seconds of measurements). To summarize, the training goal is to match the ground truth data (including air friction) with the \ac{CM} that is based on the bouncing ball \ac{FPM} (without air friction) - on a little and insufficient dataset. The goal is therefore \emph{not} to provide a perfect fit, but to perform \emph{as well as possible} under the challenging circumstances. More details about the \ac{FPM}, \ac{MLM}, the initialization of connections and on the applied training can be found in Appendix Sec. \ref{sec:setup_exp1}. 

After training, the deviations on training data are sufficiently small and play only a minor role in the validation compared to testing data. All trajectories are trained until convergence, patterns featuring the parallel structure achieve a qualitatively better fit. Training results and more details can be checked in Sec. \ref{sec:training-results}. 

\subsection{Testing}
\myfigure{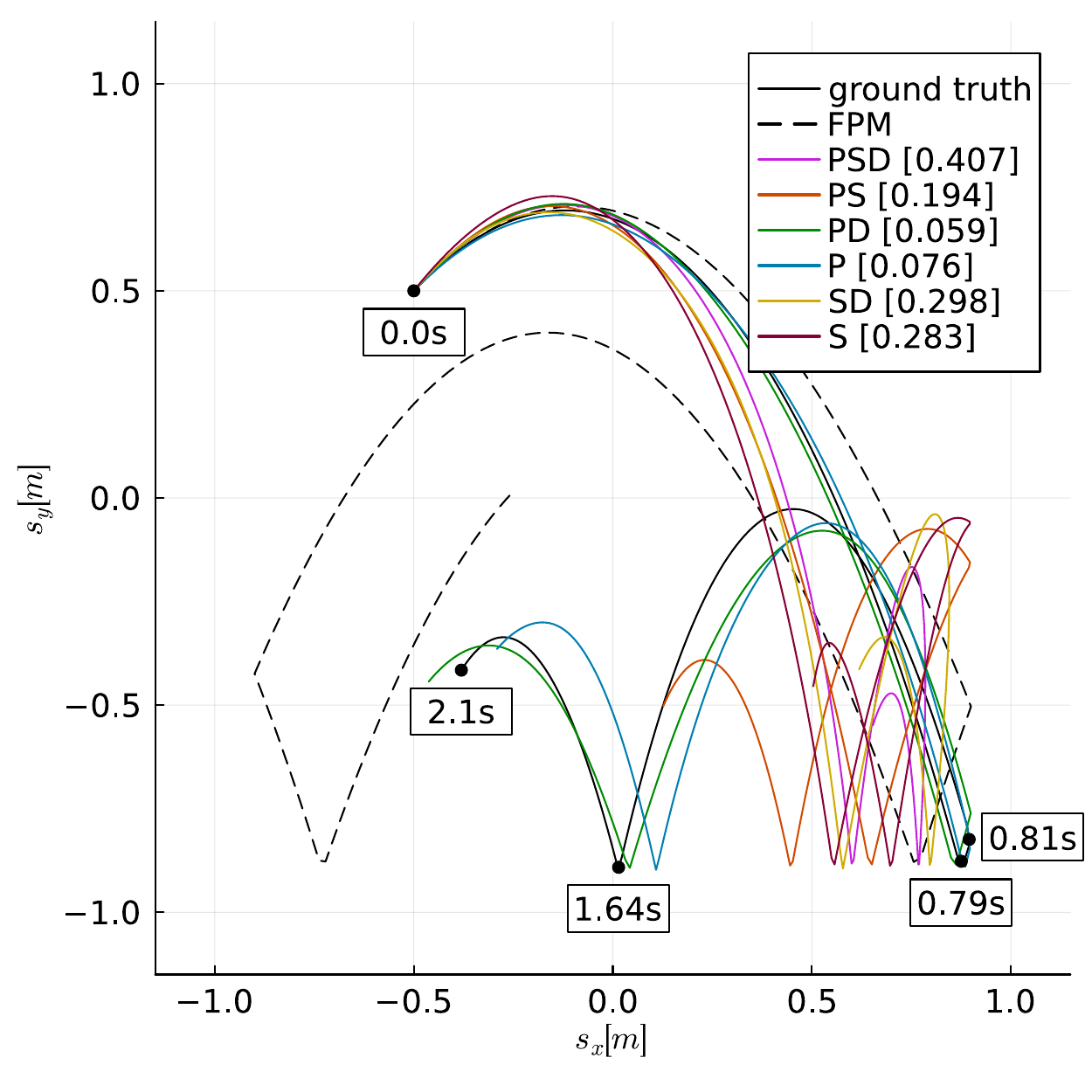}{The $s_x$-$s_y$-plane of the ball position on testing data. Loss values (compared to ground truth) are in square brackets. For testing, the ball is initialized at position $(-0.5\,m,0.5\,m)$ with velocity $(2.0\,\nicefrac{m}{s}, 2.0\,\nicefrac{m}{s})$, both are not known from training.}{exp_validation}
On testing data (compare Fig. \ref{fig:exp_validation}), we find differences compared to the training results: The more generic architectures, like \emph{PSD} and \emph{PS}, converged to poor local minima, resulting in a worse qualitative fit of the trajectories. This is explainable by a lack of data, directly leading to learning non-causal correlations. Increasing the number of different training trajectories will improve this result. On the other hand, we find that less expressive topologies, such as \emph{PD} or \emph{P}, can find better local minima even within this limited amount of data. This is the result of a better conditioned parameter space for the loss function, decreasing the degrees of freedom of the combined model by incorporation of knowledge - structural information on how the models are combined. The various aspects of such trade-offs between incorporated knowledge and the amount of data is the core topic of the field of \ac{SciML}. Because the non-parallel patterns already performed insufficiently on training data (comp. Sec. \ref{sec:training-results}), they are not investigated further on testing data. 

\subsection{Interpretation}
For easy interpretation, the absolute values within the connection matrices can be plotted in gray scale ($1$ is white, $0$ is black). The submatrix identifier and the applied scaling factor, to maximize contrast, are visible as the subplot title. 
A more detailed interpretation of the training results, comparing \emph{PSD} and \emph{P} topologies by investigating the connection matrices, can be found in the appendix \ref{sec:detailed-interpretability} and \ref{sec:appendix_exp1}.
\myfigure{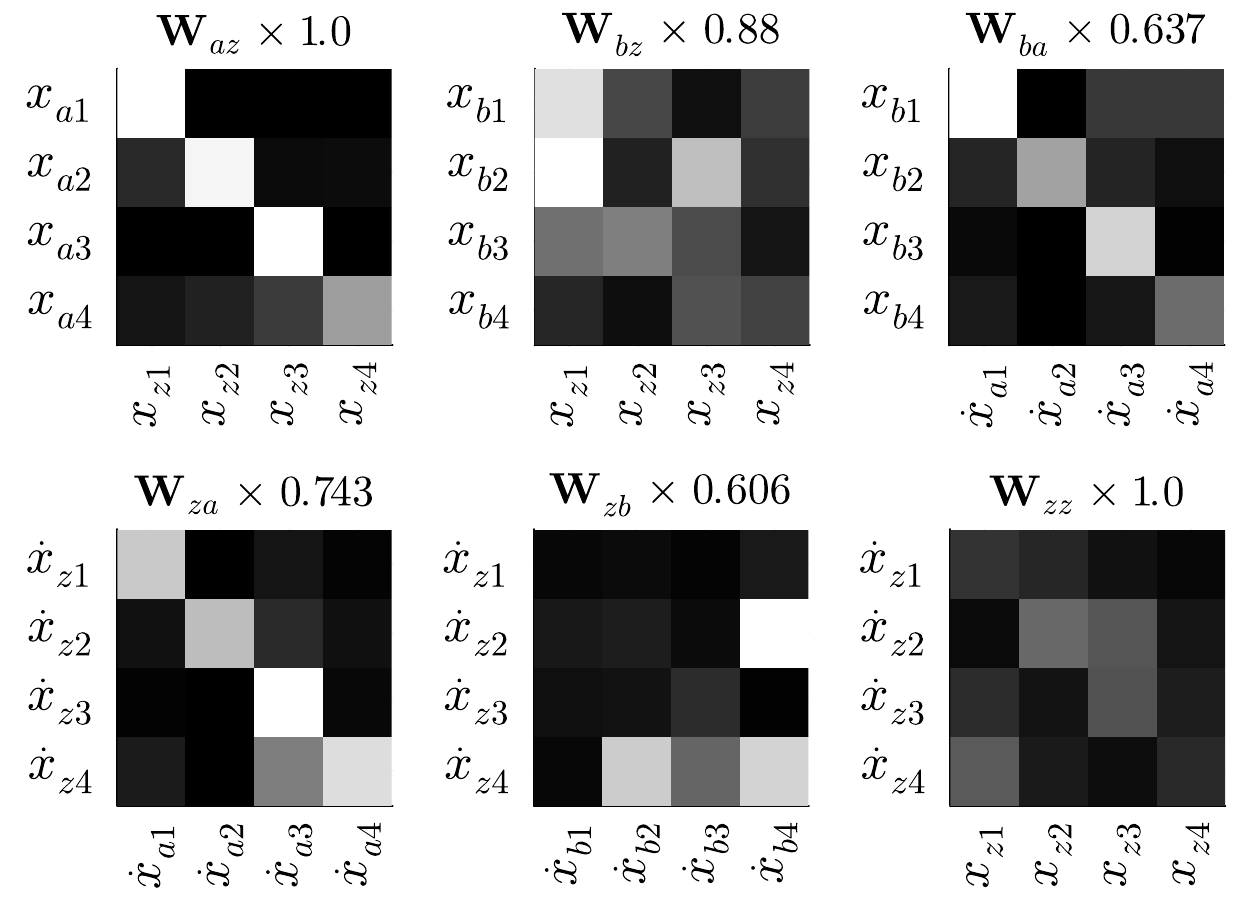}{\emph{PSD}: The matrices $\vec{W}_{az}$ and $\vec{W}_{za}$ correspond to the parallel integration of \sysa{}, the matrices $\vec{W}_{bz}$ and $\vec{W}_{zb}$ to \sysb{}, respectively. $\vec{W}_{ba}$ describes the serial connection between \sysa{} and \sysb{} and $\vec{W}_{zz}$ the \ac{DFT} between system state and derivative.}{P1S1F1O1_mod}

We know that the nonlinear friction is expressed by an additional force that is added to the balance of forces. Because this corresponds to a fraction of acceleration, we find that the parallel pattern is sufficient to add this acceleration by friction to the ball original acceleration (\ac{FPM}). This can be further investigated by comparing Eq. \ref{equ:bb-righthand} and \ref{equ:bb_air} (Appendix). Investigating Fig. \ref{fig:P1S1F1O1_mod}, we find, that the \ac{FPM} is straight-forward integrated into the \ac{CM}, which can be seen from the identity-like shapes of $\vec{W}_{az}$ (state input of \ac{FPM}) and $\vec{W}_{za}$ (derivative output of \ac{FPM}). The parallel pattern is completed by the population of $\vec{W}_{bz}$ (state input of \ac{MLM}) and $\vec{W}_{zb}$ (derivative output of \ac{MLM}). Here, $\vec{W}_{bz}$ features a linear combination of multiple signals as input for the \ac{MLM}. However, incorrect correlations are also present in the form of ball positions $x_{z1}$ and $x_{z3}$ are used as \ac{MLM} input, whereas the effect to be learned depends only on velocities. In addition, $\vec{W}_{zb}$ does not connect all \ac{MLM} outputs to the bouncing ball accelerations $\dot{x}_{z2}$ and $\dot{x}_{z4}$, as can be seen from the first column being (almost) black. Further, we also find an identity-like shape within $\vec{W}_{ba}$, indicating an unexpected serial connection in addition to the existent parallel structure. Finally, various non-causal correlations within $\vec{W}_{zz}$ should be objected. In contrast to \emph{PSD}, the topology \emph{P} restricts $\vec{W}_{ba} = \vec{W}_{zz} = \vec{0}$, making it impossible to learn the false correlations shown within \emph{PSD} and, accordingly, leads to a model with better extrapolation.

\section{Conclusion}
In this contribution, we briefly introduced a model interface, the \ac{HUDA}-\ac{ODE}, which is able to express dynamical system models from different application areas, like e.g. \ac{ML} and physics. Further, we discussed how such models can be combined to obtain a mathematical system that is solvable with or without the creation of additional algebraic loops. A special event handling routine is necessary if discontinuous models are combined and was presented. We found that the combination of \ac{HUDA}-\acp{ODE} still falls into the same class, which offers a great potential for the reuse of methods and code.

In many applications, data is sparse and rare, and optimal network architectures are unknown. It is shown that in such cases one could pick a very expressive pattern like \emph{PSD}, but wrong dependencies may be learned. Thus, while providing a method for learning arbitrary connections between submodels, using system knowledge in architecture layout is still highly desirable - if data is limited. However, if enough data is available, \emph{PSD} can be applied to explore the best architecture design.

Current and future work covers the development of new (training) methods and adaptation of existing ones, for \ac{HUDA}-\acp{ODE}. We want to show that many further state-of-the-art \ac{ML} topologies, even probabilistic models such as, e.g. (variational) auto encoders, can be expressed as \ac{HUDA}-(O)DEs including all presented advantages. Finally, it should be investigated whether the concept of \ac{HUDA}-\acp{ODE} should be extended to other types of differential equations (so \ac{HUDA}-DEs), instead of restricting it to \acp{ODE} - or if the chosen layer of abstraction is already a suitable trade-off between the relevance and expressiveness of the models to be combined and the generalizability of the corresponding methods.

%\newpage
\FloatBarrier
\section*{Abbreviations}
%\begin{multicols}{2}
	\begin{acronym}
		\acro{ANN}[ANN]{artificial neural network}
		\acro{BLT}[BLT]{block lower triangular}
		\acro{CM}[CM]{combined model}
		\acro{DFT}[DFT]{direct feed through}
		\acro{FFNN}[FFNN]{feed-forward neural network}
        \acro{FMI}[FMI]{Functional Mock-Up Interface}
		\acro{FPM}[FPM]{first principles model}
		\acro{GPU}[GPU]{graphics processing unit}
		\acro{HUDA}[HUDA]{mixed hybrid, universal, discrete, algebraic, }
		\acro{MAE}[MAE]{mean absolute error}
		\acro{ML}[ML]{machine learning}
		\acro{MLM}[MLM]{machine learning model}
		\acro{ODE}[ODE]{ordinary differential equation}
		\acro{RNN}[RNN]{recurrent neural network}
		\acro{SciML}[SciML]{Scientific Machine Learning}
		\acro{UA}[UA]{universal approximator}
		\acro{UODE}[UODE]{universal ordinary differential equation}
	\end{acronym}
%\end{multicols}

\section*{Funding}
This research was partially funded by:\\
\begin{itemize}
    \item \emph{University of Augsburg} internal funding project \emph{Forschungspotentiale besser nutzen!} (see: \url{https://www.uni-augsburg.de/de/forschung/forschungspotenziale-nutzen/}, accessed on January 13th, 2025) and
    \item ITEA4-Project OpenSCALING (Open standards for SCALable virtual engineerING and operation) No. 22013 (see: \url{https://openscaling.org/}, accessed on January 13th, 2025).
\end{itemize}
We like to thank the university leadership and ITEA for this opportunity.

\section*{Acknowledgments}
The implementation of the developed software library is highly based on many great open source libraries of the Julia programming language, some of them are: \emph{DifferentialEquations.jl} and \emph{DiffEqCallbacks.jl} \cite{DifferentialEquations.jl-2017}, \emph{SciMLSensitivity.jl} \cite{Rackauckas:2021} and \emph{ForwardDiff.jl} \cite{ForwardDiff.jl-2016}.

% \section*{Impact Statement}
% This paper presents work whose goal is to advance the field of (Scientific) Machine Learning, by providing a deeper understanding of the often casually treated topic of \emph{combining models}. As this is a very general and far-reaching topic, there are many potential societal consequences of our work, none of which we feel must be specifically highlighted here.

\printbibliography

%%%%%%%%%%%%%%%%%%%%%%%%%%%%%%%%%%%%%%%%%%%%%%%%%%%%%%%%%%%%%%%%%%%%%%%%%%%%%%%
%%%%%%%%%%%%%%%%%%%%%%%%%%%%%%%%%%%%%%%%%%%%%%%%%%%%%%%%%%%%%%%%%%%%%%%%%%%%%%%
% APPENDIX
%%%%%%%%%%%%%%%%%%%%%%%%%%%%%%%%%%%%%%%%%%%%%%%%%%%%%%%%%%%%%%%%%%%%%%%%%%%%%%%
%%%%%%%%%%%%%%%%%%%%%%%%%%%%%%%%%%%%%%%%%%%%%%%%%%%%%%%%%%%%%%%%%%%%%%%%%%%%%%%
\newpage
\appendix
\onecolumn

\section{Appendix / supplemental material}

\FloatBarrier

\subsection{Derived Topologies}
\begin{table}[ht]
	\caption{Occupancy of connection matrices for the different topologies for case (a).\\
    Legend: $\times$ = true, $\circ$ = false.}
	\label{tab:topologies}
	\centering
	\begin{tabular}{ccccccccc}
		\toprule
		P & S & D & $\vec{W}_{az} \neq \vec{0}$ & $\vec{W}_{ba} \neq \vec{0}$ & $\vec{W}_{bz} \neq \vec{0}$ & $\vec{W}_{za} \neq \vec{0}$ & $\vec{W}_{zb} \neq \vec{0}$ & $\vec{W}_{zz} \neq \vec{0}$ \\
		\midrule % \hline
		$\times$ & $\times$ & $\times$ & $\times$ & $\times$ & $\times$ & $\times$ & $\times$ & $\times$ \\
		$\times$ & $\times$ & $\circ$  & $\times$ & $\times$ & $\times$ & $\times$ & $\times$ & $\circ$  \\
		$\times$ & $\circ$  & $\times$ & $\times$ & $\circ$  & $\times$ & $\times$ & $\times$ & $\times$ \\
		$\times$ & $\circ$  & $\circ$  & $\times$ & $\circ$  & $\times$ & $\times$ & $\times$ & $\circ$  \\
		$\circ$  & $\times$ & $\times$ & $\times$ & $\times$ & $\circ$  & $\circ$  & $\times$ & $\times$ \\
		$\circ$  & $\times$ & $\circ$  & $\times$ & $\times$ & $\circ$  & $\circ$  & $\times$ & $\circ$  \\
		$\circ$  & $\circ$  & $\times$ & $\circ$  & $\circ$  & $\circ$  & $\circ$  & $\circ$  & $\times$ \\
		$\circ$  & $\circ$  & $\circ$  & $\circ$  & $\circ$  & $\circ$  & $\circ$  & $\circ$  & $\circ$  \\
		\bottomrule
	\end{tabular}
\end{table}

\FloatBarrier
\subsection{Figure and incidence matrix: \emph{PSDb}}\label{sec:appendix_caseb}
The visualization for the \emph{PSDb} architecture:
\myfigurewidth{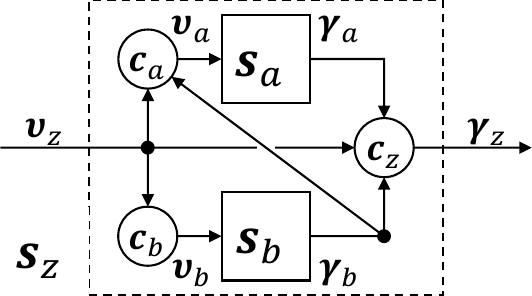}{\emph{PSDb}: $\vec{s}_b$ is evaluated before $\vec{s}_a$, resulting in $\vec{W}_{ba} = \vec{0}$.}{loop_free_b}{7cm}

The \emph{PSDb} topology features the connection matrix:
\begin{equation}
	\vec{W}_{PSDb} = 
	\kbordermatrix{
		~         & \vec{\gamma}_a    & \vec{\gamma}_b    & \vec{\upsilon}_z    \\
		\vec{c}_a & \vec{0}      & \vec{W}_{ab} & \vec{W}_{az} \\
		\vec{c}_b & \vec{0}      & \vec{0}      & \vec{W}_{bz} \\
		\vec{c}_z & \vec{W}_{za} & \vec{W}_{zb} & \vec{W}_{zz} \\
	} 
\end{equation}
For the \emph{PSDb} topology, the system incidence matrix can be transformed to \ac{BLT} form:
\begin{equation}  
	\vec{D}_{PSDb} =
	\kbordermatrix{
		~             & \vec{\gamma}_a    & \vec{\gamma}_b    & \vec{\upsilon}_a    & \vec{\upsilon}_b    & \vec{\gamma}_z \\
		\vec{s}_a     & \uvec{I}     & \vec{0}      & \vec{D}_{a}  & \vec{0}      & \vec{0}  \\
		\vec{s}_b 	  & \vec{0}      & \uvec{I}     & \vec{0}      & \vec{D}_{b}  & \vec{0}  \\
		\vec{c}_{a}   & \vec{0}      & \vec{D}_{ab} & \uvec{I}     & \vec{0}      & \vec{0}  \\
		\vec{c}_{b}   & \vec{0}      & \vec{0}      & \vec{0}      & \uvec{I}     & \vec{0}  \\
		\vec{c}_z     & \vec{D}_{za} & \vec{D}_{zb} & \vec{0}      & \vec{0}      & \uvec{I} \\
	} \eqblt
	\kbordermatrix{
		~             & \vec{\upsilon}_b    & \vec{\gamma}_b    & \vec{\upsilon}_a    & \vec{\gamma}_a    & \vec{\gamma}_z\\
		\vec{c}_{b}   & \uvec{I}     & \vec{0}      & \vec{0}      & \vec{0}      & \vec{0}  \\
		\vec{s}_b 	  & \vec{D}_{b}  & \uvec{I}     & \vec{0}      & \vec{0}      & \vec{0}  \\
		\vec{c}_{a}   & \vec{0}      & \vec{D}_{ab} & \uvec{I}     & \vec{0}      & \vec{0}  \\
		\vec{s}_a     & \vec{0}      & \vec{0}      & \vec{D}_{a}  & \uvec{I}     & \vec{0}  \\
		\vec{c}_z     & \vec{0}      & \vec{D}_{zb} & \vec{0}      & \vec{D}_{za} & \uvec{I} \\
	} 
\end{equation}

\FloatBarrier
\subsection{Event in subsystem \sysa{} for the \emph{PSDa} topology}\label{sec:appendix_eventa}
For the example of an event in the subsystem \sysa{} (for the topology \emph{PSDa}), the incidence matrix of the system at event $\vec{D}^{(a)}_{PSDa}$ states:
\begin{equation}
	\vec{D}^{(a)}_{PSDa} \eqblt
	\kbordermatrix{
		~             & \vec{\gamma}_a    & \vec{\upsilon}_z    & \vec{\upsilon}_b    & \vec{\gamma}_b    & \vec{\gamma}_z\\
		\vec{s}_a     & \uvec{I}     & \vec{0}      & \vec{0}      & \vec{0}      & \vec{0}  \\
		\vec{c}_{a}   & \vec{0}      & \uvec{D}_{az}& \vec{0}      & \vec{0}      & \vec{0}  \\
		\vec{c}_{b}   & \vec{D}_{ba} & \vec{D}_{bz} & \uvec{I}     & \vec{0}      & \vec{0}  \\
		\vec{s}_b 	  & \vec{0}      & \vec{0}      & \vec{D}_{b}  & \uvec{I}     & \vec{0}  \\
		\vec{c}_{z}   & \vec{D}_{za} & \vec{D}_{zz} & \vec{0}      & \vec{D}_{zb} & \uvec{I} \\
	} 
    .
\end{equation}
Comparing to the system of equations without event (s. equation \ref{equ:D_PSD}), only the first two rows are switched. The residual for the case that $\vec{W}_{az}$ is not invertible states:
\begin{equation} 
	\vec{r}^{(a)}(\vec{\upsilon}_z) = 
	\vec{W}_{az} \vec{\upsilon}_z -
	\vec{\upsilon}_{a}
    .
\end{equation}
If $\vec{W}_{az}$ is invertible (for example if it is a permutation matrix), the analytical solutions can be derived for the global state $\vec{\upsilon}_z$ directly:
\begin{equation} 
	\vec{\upsilon}_z = 
	\vec{W}_{az}^{-1} \vec{\upsilon}_a
    .
\end{equation}

\subsection{Remark on the choice of experiments}\label{sec:remark}
In the field of machine learning, methods are often derived based on a hypothesis and then validated by experiments. In the fundamental subject area presented, validation by experiment is challenging because, regardless of the size of the experiment, proofs and statements are only ever made for the small subarea that is investigated. We therefore pursue the goal of generating the plausibility and validity of the method via the actual mathematical derivation in Sec. \ref{sec:model} and \ref{sec:paradigm} instead of solely by experiment. However, we don't want to use this as an excuse for not \emph{trying} to provide suitable experiments and further added an experiment comparing the combination of some \ac{ML} architecture in the Appendix \ref{sec:common-interface} - knowing that this only covers a small subset of the investigated domain. In summary, the experiments within this paper are not intended to be interpreted as a general validation of the presented method for the entire application domain, but rather to shed light on the applicability and benefits of it.

\subsection{Experimental setup}\label{sec:setup_exp1}
All experiments were performed in the Julia programming language \cite{Bezanson:2017}. The experiment code to reproduce the results can be found in the library repository\footnote{\githuburl{}}. In addition to Julia, the proposed combination topologies are also implementable within different acausal modeling languages like \emph{Modelica} \cite{Modelica:2024} and modeling/programming tools like \emph{ModelingToolkit.jl} \cite{Ma:2021}, as well as causal modeling languages like \emph{Simulink} \cite{Simulink:2024}. However, most modeling tools\footnote{This explicitly excludes \emph{ModelingToolkit.jl}, which supports automatic differentiation.} are still missing crucial feature implementations to allow for \emph{learnable} connections (for example automatic differentiation).

The goal of the \ac{CM} is to predict the more detailed simulation, including air friction, based on the less detailed \ac{FPM} and the \ac{MLM} (\ac{ANN}) - but on insufficient training data.

\paragraph{Data and loss function}
Data is provided by a more detailed computer simulation, which further models air friction of the ball (quadratic speed-dependent force; equations are given below). Four scenarios are used for training, with four different start positions and velocities for the bouncing ball. As loss function, the well known \ac{MAE} between the \ac{CM} states $s_x$ and $s_y$ (the ball position, normalized by scaling with $0.5$) and $v_x$ and $v_y$ (the ball velocity, normalized by scaling with $0.1$) and the corresponding ground truth data is deployed. For every gradient determination, one of the four scenarios is randomly picked.

\paragraph{Training}
The \ac{CM} is optimized by the Adam \cite{Kingma:2017} optimizer (default parameterization, $\eta=10^{-3}$) with \emph{growing horizon} training strategy, meaning the simulation trajectory is limited to the first 5\% (over time) and successively increased by $0.01\,s$, if the current \ac{MAE} of the worst of the training scenarios falls under $0.05$. We perform 20,000 training steps, which is enough so that all topologies reach the full training horizon and converge to a local minimum. 

\paragraph{\Acf{MLM}}
The \ac{MLM} is a fully-connected \ac{FFNN} with two layers of sizes $(4,8)$ and $(8,2)$ and hyperbolic tangent activation for both layers. The linear connection between \ac{ANN} and \ac{CM} allows for an unbounded \ac{ANN} output, as it can be interpreted as an additional linear layer. Note that because of the second order \ac{ODE}, only ball accelerations are used as \ac{ANN} outputs, and velocities are implemented as \ac{DFT}, because they are part of the state of the system, as well as the state derivative.

\paragraph{\Acf{FPM}}
The \ac{FPM} is described by the \ac{ODE} of a bouncing ball in two dimensions, with walls at $x=\{-1,1\}$ and $y =\{-1,1\}$. Every time the ball hits a vertical (or horizontal) wall, the ball's velocity in direction $x$ (or $y$) is inverted and reduced by a constant fraction (energy dissipation). The state vector of the \ac{FPM} (as well as of the \ac{CM}) notes
\begin{equation}
	\vec{x} = 
	\begin{bmatrix}
		s_x \\ 
		v_x \\
		s_y \\
		v_y
	\end{bmatrix}
    .
\end{equation}
The simulation is started in one of the scenarios $s$ with start state
\begin{equation}
	\vec{x}_0 = \vec{x}_{c0} \in
	\begin{cases}
		[-0.25, -6.0, 0.5, 8.0]^T & \text{if } s=1\\
		[0.5, 8.0, -0.25, -6.0]^T & \text{if } s=2\\
        [0.0, 0.0, 0.0, 4.0, 0.0]^T & \text{if } s=3\\
		[-0.75, 4.0, -0.5, 6.0]^T & \text{if } s=4\\
        [-0.5, 2.0, 0.5, 2.0, 0.0]^T & \text{if } s=5
	\end{cases}
    \quad
    ,
\end{equation} 
where $s$ is the scenario identifier. Scenarios $s \in \{1, 2, 3, 4\}$ are used for training, whereas scenario $s=5$ is used for testing. The system is solved for $t \in [0.0, 2.1]$. The system does not feature an input, therefore $\vec{u}$ is omitted for readability in the following. The dynamics of the system are defined by the function $\vec{f}_c$ as follows:
\begin{equation}\label{equ:bb-righthand}
	\vec{f}_c(\vec{x}, \vec{p}, t) =
	\begin{bmatrix}
		v_x \\ 
		0 \\
		v_y \\
		-g
	\end{bmatrix}
    ,
\end{equation}
with $g \approx 9.81$ the gravity constant. The model for generation of ground-truth data features a different right-hand side including air friction and states
\begin{equation}\label{equ:bb_air}
	\vec{f}_{gt}(\vec{x}, \vec{p}, t) =
	\begin{bmatrix}
		v_x \\ 
		0  - v_x \cdot v \cdot \mu / m\\
		v_y \\
		-g - v_y \cdot v \cdot \mu / m
	\end{bmatrix}
    \quad \text{where} \quad v = \sqrt{v_x^2 + v_y^2}
    ,
\end{equation}
with air friction coefficient $\mu = 0.15$. The event condition function is defined by the distance between the ball and the four walls at $x=\{-1,1\}$ and $y =\{-1,1\}$:
\begin{equation}\label{equ:bb_c}
	\vec{c}(\vec{x}, \vec{p}, t) =
	\begin{bmatrix}
		1+s_x-r \\
		1-s_x-r \\
		1+s_y-r \\
		1-s_y-r 
	\end{bmatrix}
    ,
\end{equation}
with $r = 0.1$ the ball radius. The corresponding event affect function states
\begin{equation}\label{equ:bb_a}
    \vec{x}^+ = 
	\vec{a}(\vec{x}^-, \vec{p}, t^-, \vec{q}) =
	\begin{cases}
		\begin{bmatrix}
			-1+r \\
			-v_x^- \cdot (1-d) \\
			s_y^- \\
			v_y^-
		\end{bmatrix} & \text{if } q_1=1\\
		\begin{bmatrix}
			1-r \\
			-v_x^- \cdot (1-d) \\
			s_y^- \\
			v_y^-
		\end{bmatrix} & \text{if } q_2=1\\
		\begin{bmatrix}
			s_x^- \\
			v_x^- \\
			-1+r \\
			-v_y^- \cdot (1-d)
		\end{bmatrix} & \text{if } q_3=1\\
		\begin{bmatrix}
			s_x^- \\
			v_x^- \\
			1-r \\
			-v_y^- \cdot (1-d)
		\end{bmatrix} & \text{if } q_4=1\\
	\end{cases}
    \quad 
    ,
\end{equation}
with energy dissipation factor $d=0.1$. The event identifiers $q_i \in \{0,1\}$ for $i \in \{1,2,3,4\}$ distinguish between the four different collisions (left, right, bottom, top).

\paragraph{Initialization of connections}
All submatrices of the connection matrix $\vec{W}$ are initialized and optimized or set to zero according to the chosen topology. The bias $\vec{b}$ is not used (zeros) within the experiment, which allows for a more interesting discussion. A more detailed description of the initialization used for the connections is given in table \ref{tab:exp1}. Biases are initialized with zeros in any case.
\begin{table}[ht]
	\caption{Connection matrix $\vec{W}$ initialization. Legend: $\cross$ = true, $\circ$ = false, $\vec{I}$ = identity (static), $\hvec{I}$ = identity (noisy, trainable), $\vec{0}$ = zero matrix (static), $\hvec{0}$ = zero matrix (noisy, trainable).}
	\label{tab:exp1}
	\centering
	\begin{tabularx}{\linewidth}{ X X X X X X X X X X X X }
		\toprule
		S & P & D & $\vec{W}_{az}$ & $\vec{W}_{ba}$ & $\vec{W}_{bz}$ & $\vec{W}_{za}$ & $\vec{W}_{zb}$ & $\vec{W}_{zz}$ \\
		\midrule % \hline
		$\times$ & $\times$ & $\times$ & $\hvec{I}$ & $\hvec{I}$ & $\hvec{0}$ & $\hvec{I}$ & $\hvec{0}$ & $\hvec{0}$ \\
		$\circ$  & $\times$ & $\times$ & $\hvec{I}$ & $\vec{0}$  & $\hvec{0}$ & $\hvec{I}$ & $\hvec{0}$ & $\hvec{0}$ \\
		$\times$ & $\circ$  & $\times$ & $\hvec{I}$ & $\hvec{I}$ & $\vec{0}$  & $\vec{0}$  & $\hvec{I}$ & $\hvec{0}$ \\
		$\circ$  & $\circ$  & $\times$ & $\vec{0}$  & $\vec{0}$  & $\vec{0}$  & $\vec{0}$  & $\vec{0}$  & $\hvec{0}$ \\
		$\times$ & $\times$ & $\circ$  & $\hvec{I}$ & $\hvec{I}$ & $\hvec{0}$ & $\hvec{I}$ & $\hvec{0}$ & $\vec{0}$ \\
		$\circ$  & $\times$ & $\circ$  & $\hvec{I}$ & $\vec{0}$  & $\hvec{0}$ & $\hvec{I}$ & $\hvec{0}$ & $\vec{0}$ \\
		$\times$ & $\circ$  & $\circ$  & $\hvec{I}$ & $\hvec{I}$ & $\vec{0}$  & $\vec{0}$  & $\hvec{I}$ & $\vec{0}$ \\
		$\circ$  & $\circ$  & $\circ$  & $\vec{0}$  & $\vec{0}$  & $\vec{0}$  & $\vec{0}$  & $\vec{0}$  & $\vec{0}$ \\
		\bottomrule
	\end{tabularx}
\end{table}

\FloatBarrier
\subsection{Training Results}\label{sec:training-results}

\begin{figure}[ht]
    \centering
    \begin{subfigure}[b]{0.475\textwidth}  
        \centering
        \includegraphics[width=\textwidth]{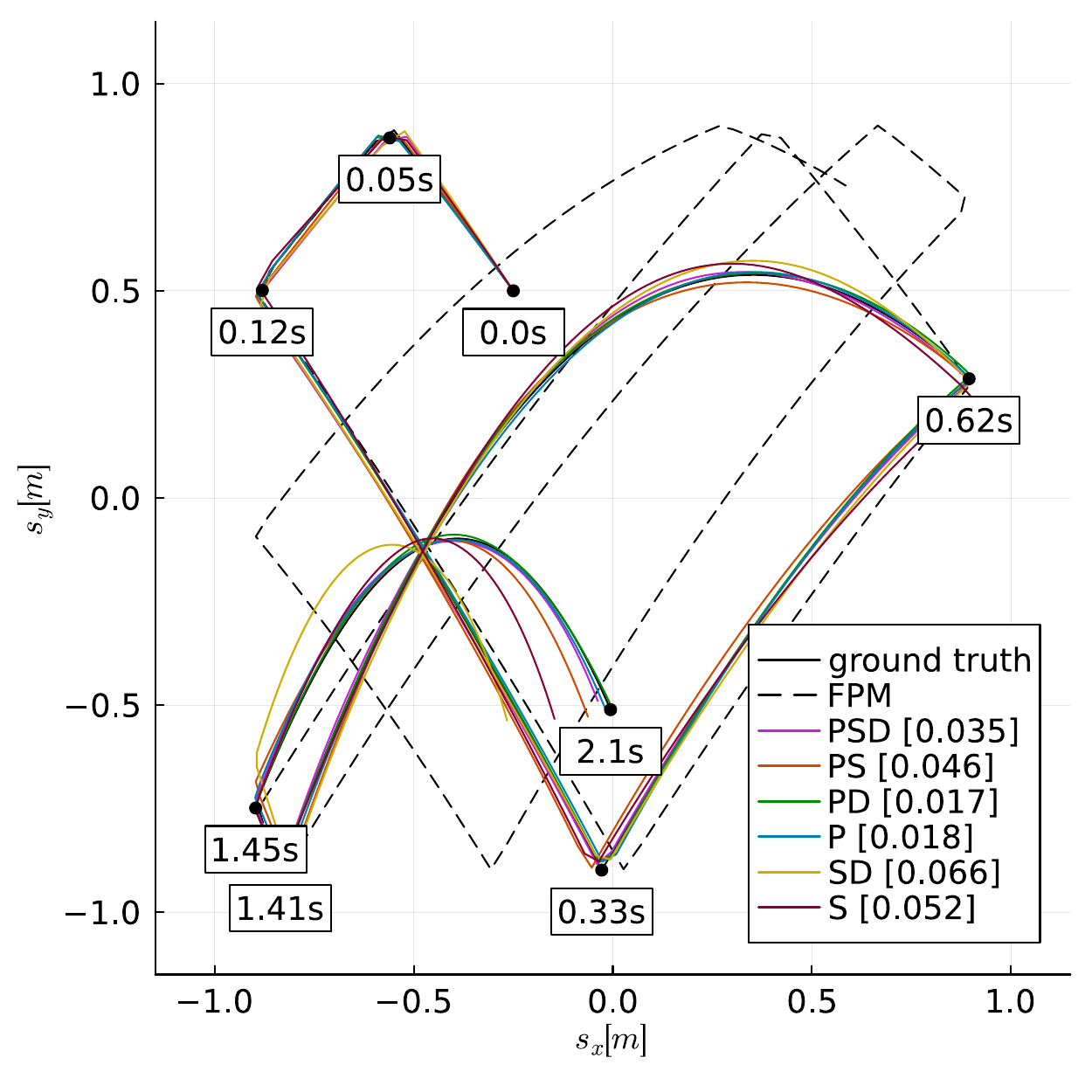}
        \caption{$s=1$}
    \end{subfigure}
    \hfill
    \begin{subfigure}[b]{0.475\textwidth}  
        \centering 
        \includegraphics[width=\textwidth]{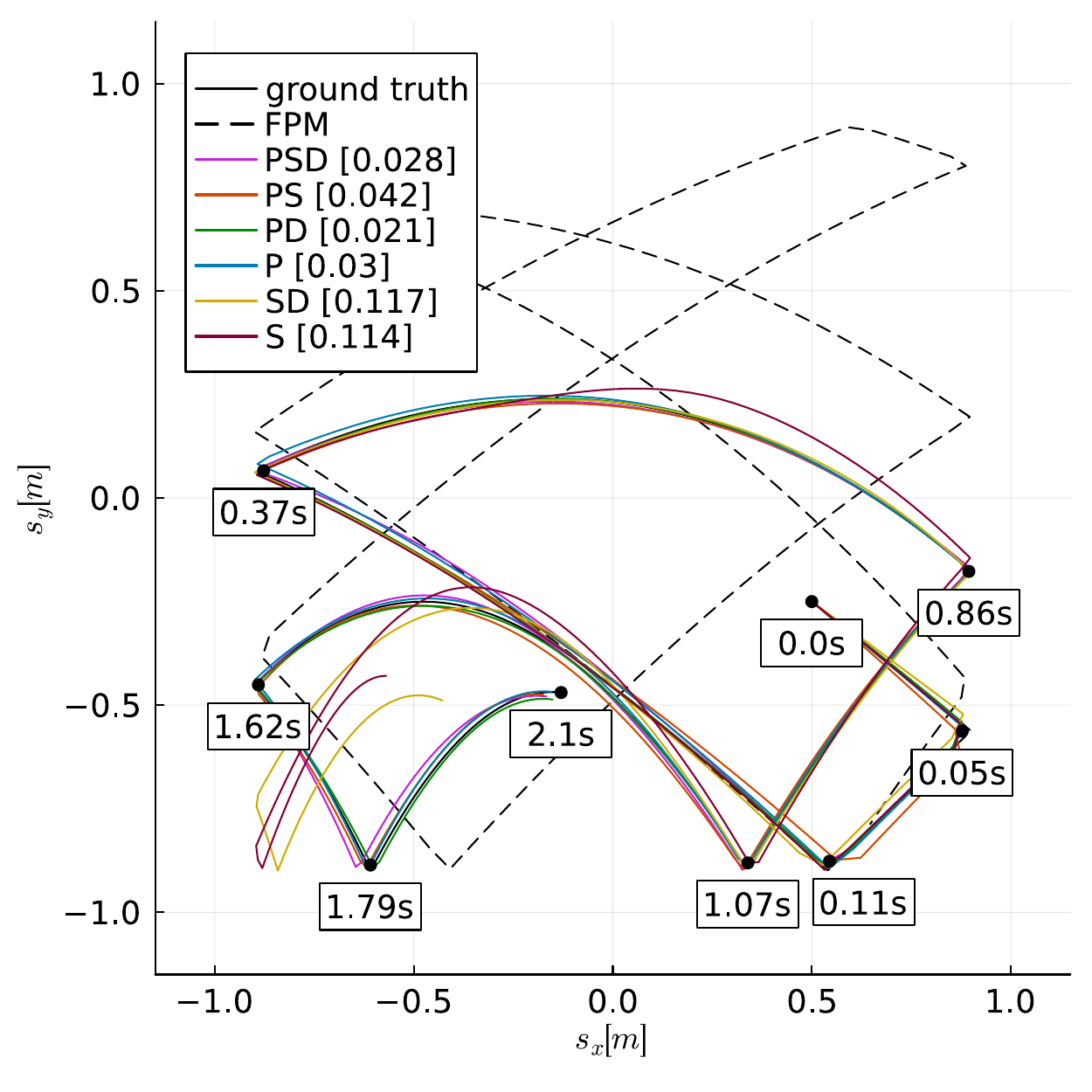}
        \caption{$s=2$}
    \end{subfigure}
    \vskip\baselineskip
    \begin{subfigure}[b]{0.475\textwidth} 
        \centering 
        \includegraphics[width=\textwidth]{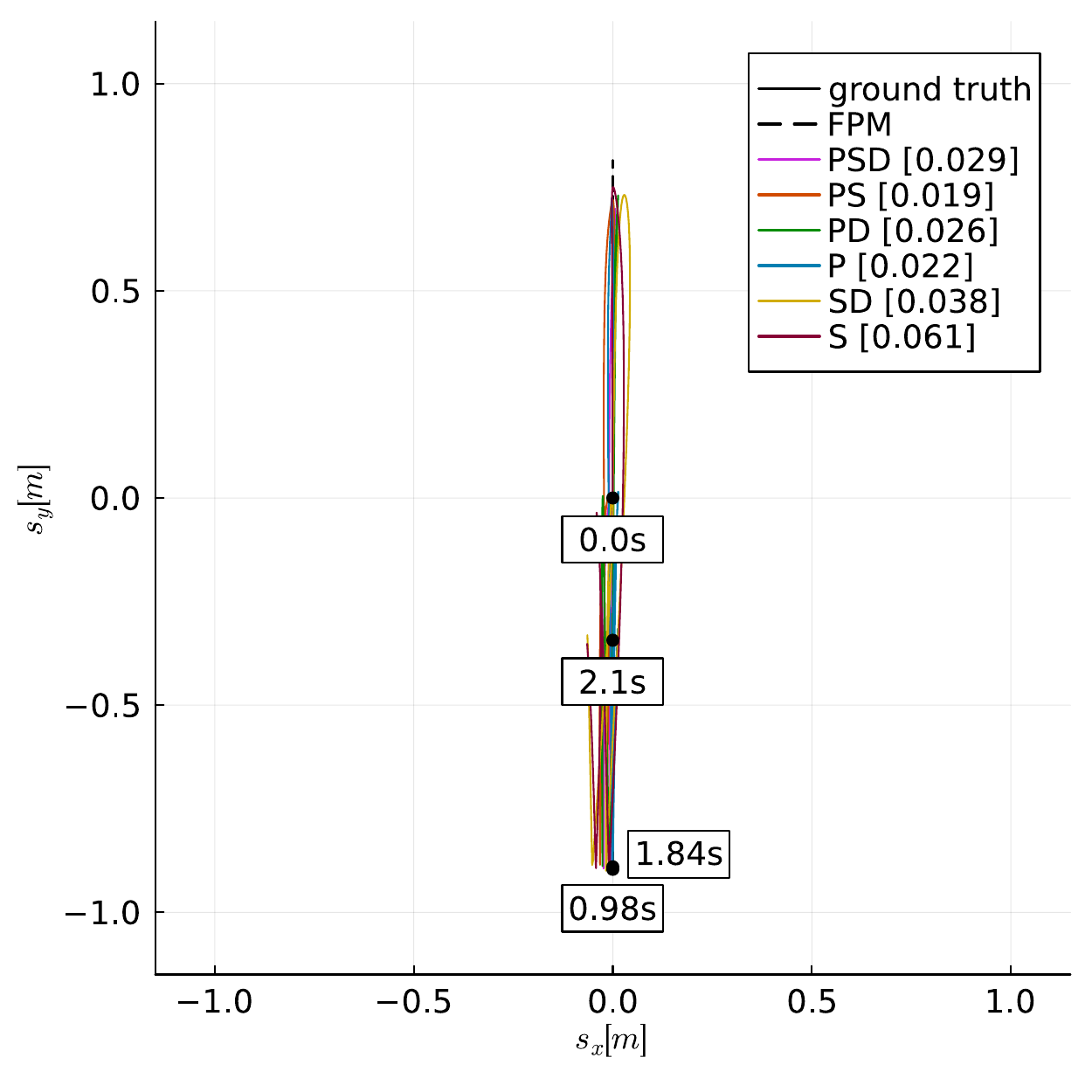}
        \caption{$s=3$}
    \end{subfigure}
    \hfill
    \begin{subfigure}[b]{0.475\textwidth}   
        \centering 
        \includegraphics[width=\textwidth]{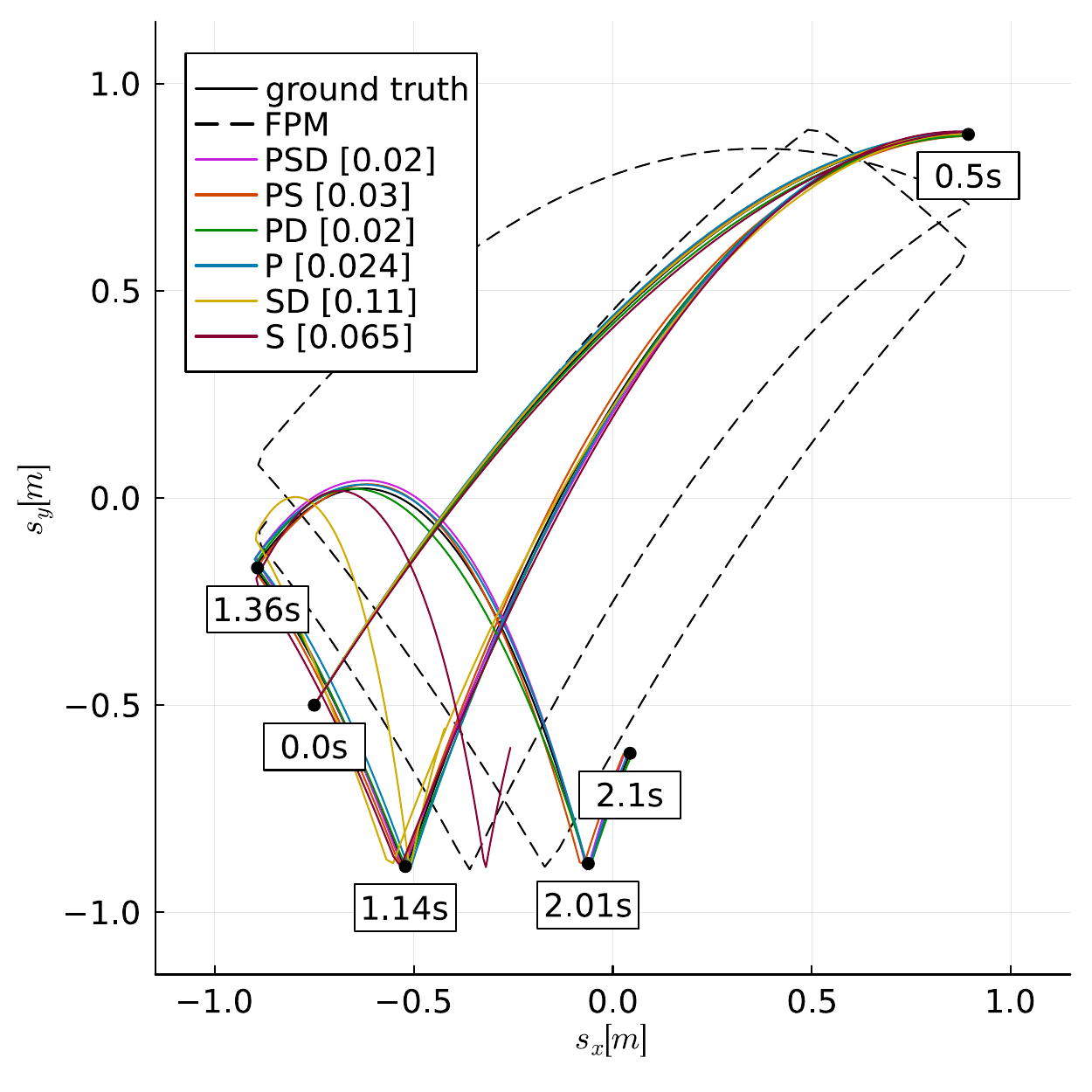}
        \caption{$s=4$}
    \end{subfigure}
    \caption{Training results of the first experiment: The $s_x$-$s_y$-plane of the position of the ball on training data. Loss values (compared to ground truth) are in square brackets. Points in time for start, stop and collisions are marked with black dots and time labels, to track the simulation trajectory over time.} 
    \label{fig:exp_topologies}
\end{figure}
    
Investigating the learned trajectories of the bouncing ball in Fig. \ref{fig:exp_topologies}, the black (dashed) plot shows the simulation trajectory of the \ac{FPM} (without air friction). We can observe a large deviation between the model including air friction (black, solid, covered by the colored trajectories) and without air friction (black, dashed). 

If we check for the results of the topologies \emph{PSD}, \emph{PS}, \emph{PD} and \emph{P} (different colors) all match the ground truth trajectory (black, covered), and we for now assume that the friction effect is learned in terms of a qualitative analysis. The topologies without parallel patterns, \emph{SD} and \emph{S}, have difficulties to match the data. This is because of the missing bias $\vec{b}$, which is necessary to compensate for the constant acceleration in $y$ by gravity, compare Eq. \ref{equ:bb-righthand} (appendix) for further details. As a side note, without proof, these two patterns also converge to better minima if the trainable bias is applied. 
% Rating the general trend, investigating the loss values, we find that less expressive patterns (like \emph{PD} or \emph{P}) converge to better local minima compared to the more expressive (like \emph{PSD} or \emph{PS}) patterns - even on training data.

\FloatBarrier
\subsection{Detailed interpretation}\label{sec:detailed-interpretability}
\myfiguremode{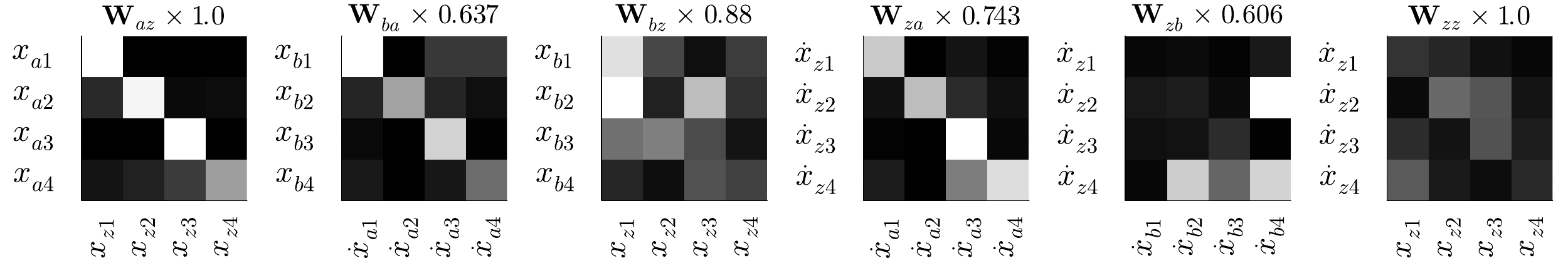}{\emph{PSD}: The matrices $\vec{W}_{az}$, $\vec{W}_{ba}$, and $\vec{W}_{za}$ are initialized with (noisy) identity, the remaining with (noisy) zeros. All matrix weights are optimized together with the \ac{ANN} parameters.}{P1S1F1O1}{14cm}{H}
\myfiguremode{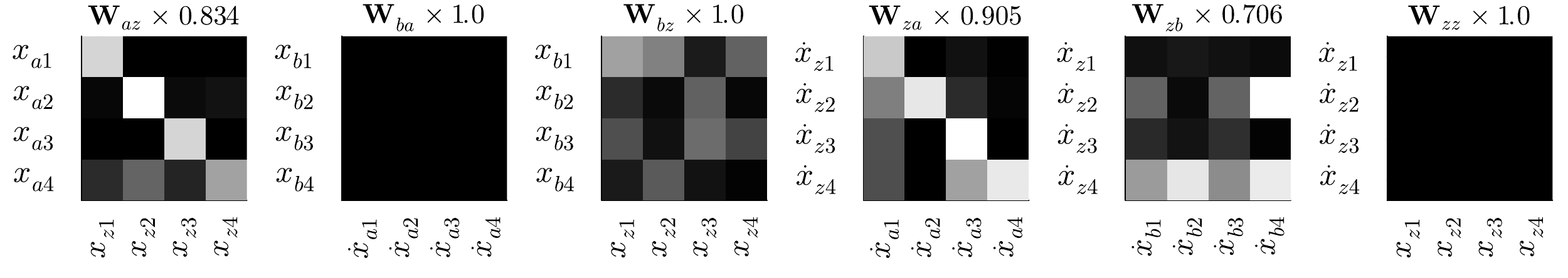}{\emph{P}: The matrices $\vec{W}_{az}$, $\vec{W}_{ba}$, and $\vec{W}_{za}$ are initialized with (noisy) identity, the remaining with (noisy) zeros. All matrix weights except $\vec{W}_{zz}$ are optimized together with the \ac{ANN} parameters.}{P1S0F0O1}{14cm}{H}

Detailed interpretation and explanation of the training results of the \emph{PSD} and \emph{P} topologies are given in the following, also discussing why \emph{P} performs better than \emph{PSD} on testing data:
\begin{description}
	\item[$\vec{W}_{az}$] The matrix visualizes the connection between the global system state $\vec{x}_z$ (input of \ac{CM}) and subsystem \sysa{} state $\vec{x}_a$ (input of \ac{FPM}). For both topologies, it is initialized with (noisy) identity and almost remains in its identity shape, showing a strict connection between the global state and the local state of the \ac{FPM}.
	\item[$\vec{W}_{ba}$] The matrix visualizes the connection between the subsystem \sysa{} local state derivative $\dvec{x}_a$ (output of \ac{FPM}) and subsystem \sysb{} local state $\vec{x}_b$ (input of \ac{MLM}). This corresponds to the serial connection between the \ac{FPM} and \ac{MLM}. For \emph{PSDa}, it is initialized with (noisy) identity and, like $\vec{W}_{az}$, it stays close to its identity shape after training. This results in a strong serial connection between the subsystem \sysa{} (output) and \sysb{} (input). For \emph{P}, the part of the matrix is set to zeros and is not trained (no serial connection).
	\item[$\vec{W}_{bz}$] The matrix visualizes the connection between the global system state $\vec{x}_z$ (input of \ac{CM}) and subsystem \sysb{} local state $\vec{x}_b$ (input of \ac{MLM}). It is initialized with (noisy) zeros. After training, a linear combination of states is used as input to the \ac{MLM} for both topologies. Note, that also the ball positions (first and third column, $x_{z1}$ and $x_{z3}$) are connected to the \ac{MLM}, which is not expected because the friction effect only depends on the velocities. Again, because of lack of data, wrong correlations are present.
	\item[$\vec{W}_{za}$] The matrix visualizes the connection between the subsystem \sysa{} state derivative $\dvec{x}_a$ (output of \ac{FPM}) and global state derivative $\dvec{x}_z$ (output of \ac{CM}). It is initialized with an (noisy) identity and remains in an identity-like shape. There are small non-zero entries for \emph{P}, indicating that a mix of signals is used. However, this is not an issue because the impact on the \ac{CM} state derivative can be shared between \ac{FPM} and \ac{MLM}. The overall structure of the matrix indicates a strong connection between \ac{FPM} (output) and the global output of the system, resulting (together with $\vec{W}_{zb}$) in a parallel structure.
	\item[$\vec{W}_{zb}$] The matrix visualizes the connection between the subsystem \sysb{} state derivative $\dvec{x}_b$ (output of \ac{MLM}) and global state derivative $\dvec{x}_z$ (output of \ac{CM}). It is initialized with (noisy) zeros. For the pattern \emph{P}, the \ac{MLM} outputs are connected to the ball accelerations of the \ac{CM} (second and fourth row, $\dot{x}_{z2}$ and $\dot{x}_{z4}$). The topology \emph{PSD} on the other hand, does not use the entire \ac{MLM} output, as can be seen from the first zero column.
	\item[$\vec{W}_{zz}$] The matrix visualizes the connection between the state of the \ac{CM} input $\vec{x}_z$ (state) and \ac{CM} output  $\dvec{x}_z$ (state derivative). For \emph{PSD}, it is initialized with (noisy) zeros and some unexpected connections can be found here for the \emph{PSD} topology, based on noncausal correlations in the very small data set. Because \emph{P} does not use this matrix (zeros), it has no access to the wrong correlations.
    % The strongest ones are at $(2,3)$ and $(2,1)$ (the connection between ball positions and the ball acceleration in $y$), which is obviously non-causal, but can be explained by the correlation anaylysis in the appendix, see figure \ref{fig:cc}.  
\end{description}

\subsection{Additional experimental results}\label{sec:appendix_exp1}
Results from the other topologies can be found below. 
\myfiguremode{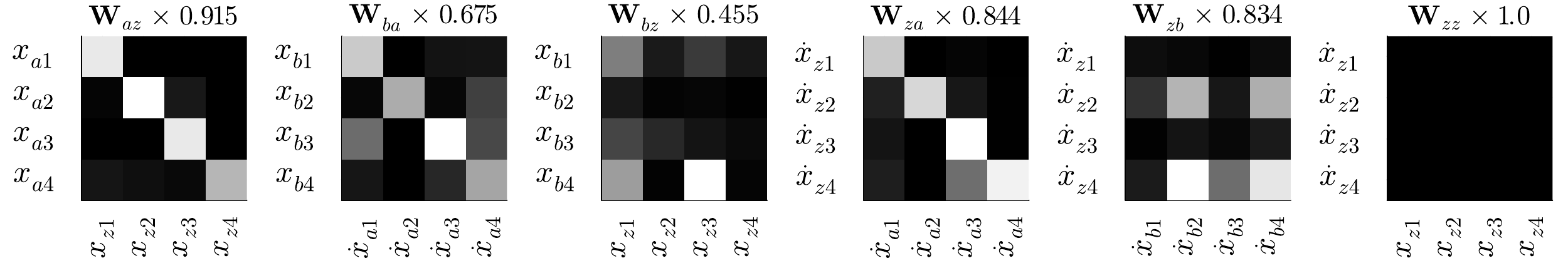}{\emph{PS}: The matrices $\vec{W}_{az}$, $\vec{W}_{ba}$, and $\vec{W}_{za}$ are initialized with (noisy) identity, the remaining with (noisy) zeros. All matrix weights except $\vec{W}_{zz}$ are optimized together with the \ac{ANN} parameters.}{P1S1F0O1}{14cm}{H}
\myfiguremode{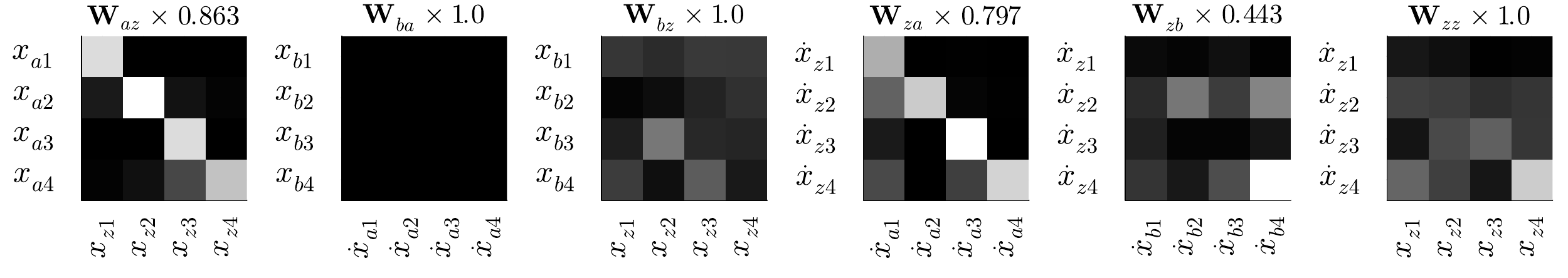}{\emph{PD}: The matrices $\vec{W}_{az}$ and $\vec{W}_{za}$ are initialized with (noisy) identity, the remaining with (noisy) zeros. All matrix weights except $\vec{W}_{ba}$ are optimized together with the \ac{ANN} parameters.}{P1S0F1O1}{14cm}{H}
\myfiguremode{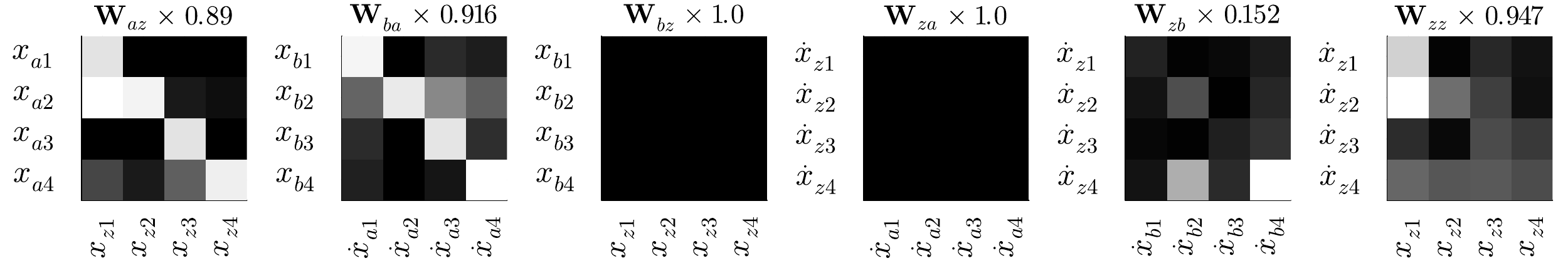}{\emph{SD}: The matrices $\vec{W}_{az}$ and $\vec{W}_{ba}$ are initialized with (noisy) identity, the remaining with (noisy) zeros. All matrix weights except $\vec{W}_{bz}$ and $\vec{W}_{za}$ are optimized together with the \ac{ANN} parameters.}{P0S1F1O1}{14cm}{H}
\myfiguremode{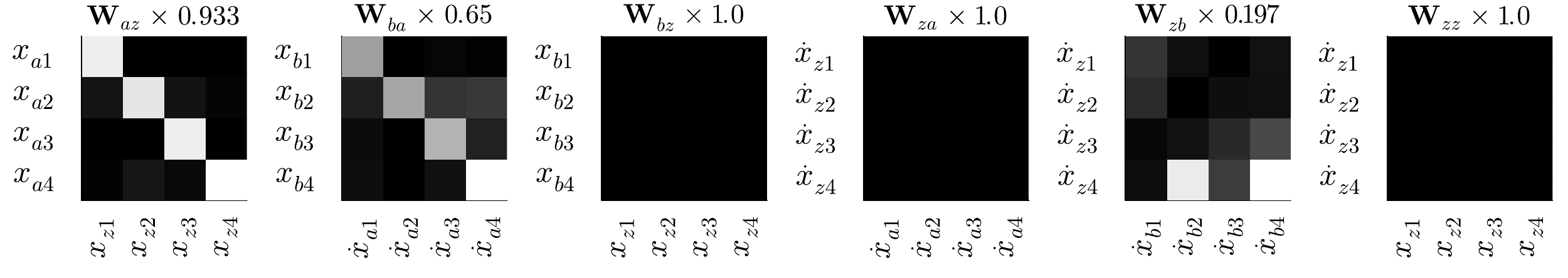}{\emph{S}: The matrices $\vec{W}_{az}$ and $\vec{W}_{ba}$ are initialized with (noisy) identity, the remaining with (noisy) zeros. All matrix weights except $\vec{W}_{bz}$, $\vec{W}_{za}$ and $\vec{W}_{zz}$ are optimized together with the \ac{ANN} parameters.}{P0S1F0O1}{14cm}{H}

\FloatBarrier
\subsection{Experiment: Common interface}\label{sec:common-interface}
In this second experiment, we want to show that four different \acp{MLM} can be expressed in form of \ac{HUDA}-\acp{ODE} and be trained using the very same algorithm without any adaptions. Again, this is not intended to assess accuracy, but to show that very different models can be expressed as \ac{HUDA}-\acp{ODE} and trained without any methodological adaptations. We compare the following model classes: 
\begin{description}
    \item [Continuous]: A neural \ac{ODE} \cite{Chen:2018} is implemented by inference of a \ac{FFNN} within the function $\vec{f}_c$ of the \ac{HUDA}-\ac{ODE}. In this case, the system state derivative is computed by the \ac{FFNN} based on the system state. No further function implementations are required.
    \item [Discrete]: A \ac{RNN} is implemented by defining the \ac{RNN} hidden state as the discrete state within the \ac{HUDA}-\ac{ODE}. This discrete state is only updated at predefined time events, which can be interpreted as the \emph{sampling frequency} (here $10\,Hz$) of the \ac{RNN}. This requires the implementation of functions $\vec{a}$ and $\vec{c}$. The event condition $\vec{c}$ is implemented so that after every sampling period, a time event is triggered. In case of an event, therefore within the function $\vec{a}$, the next discrete state is evaluated by inference of a \ac{FFNN} based on the previous discrete state.
    \item [Continuous + Event] Based on the \emph{continuous} example, the neural \ac{ODE} can further be extended by event functions $\vec{a}$ and $\vec{c}$, implementing the bouncing ball state events (compare Eq. \ref{equ:bb_c} and \ref{equ:bb_a}).
    \item [Discrete + Event] In analogy, the \ac{RNN} within the \emph{discrete} example can be extended by the bouncing ball state events.
\end{description}
Fig. \ref{fig:common} compares the four models. The bouncing ball from the previous example was reused for data generation. All models are trained under the same conditions; however, a very easy trajectory needs to be applied, to allow all models (especially the simple \ac{RNN}) to converge without changing hyperparameters.

\myfigurewidth{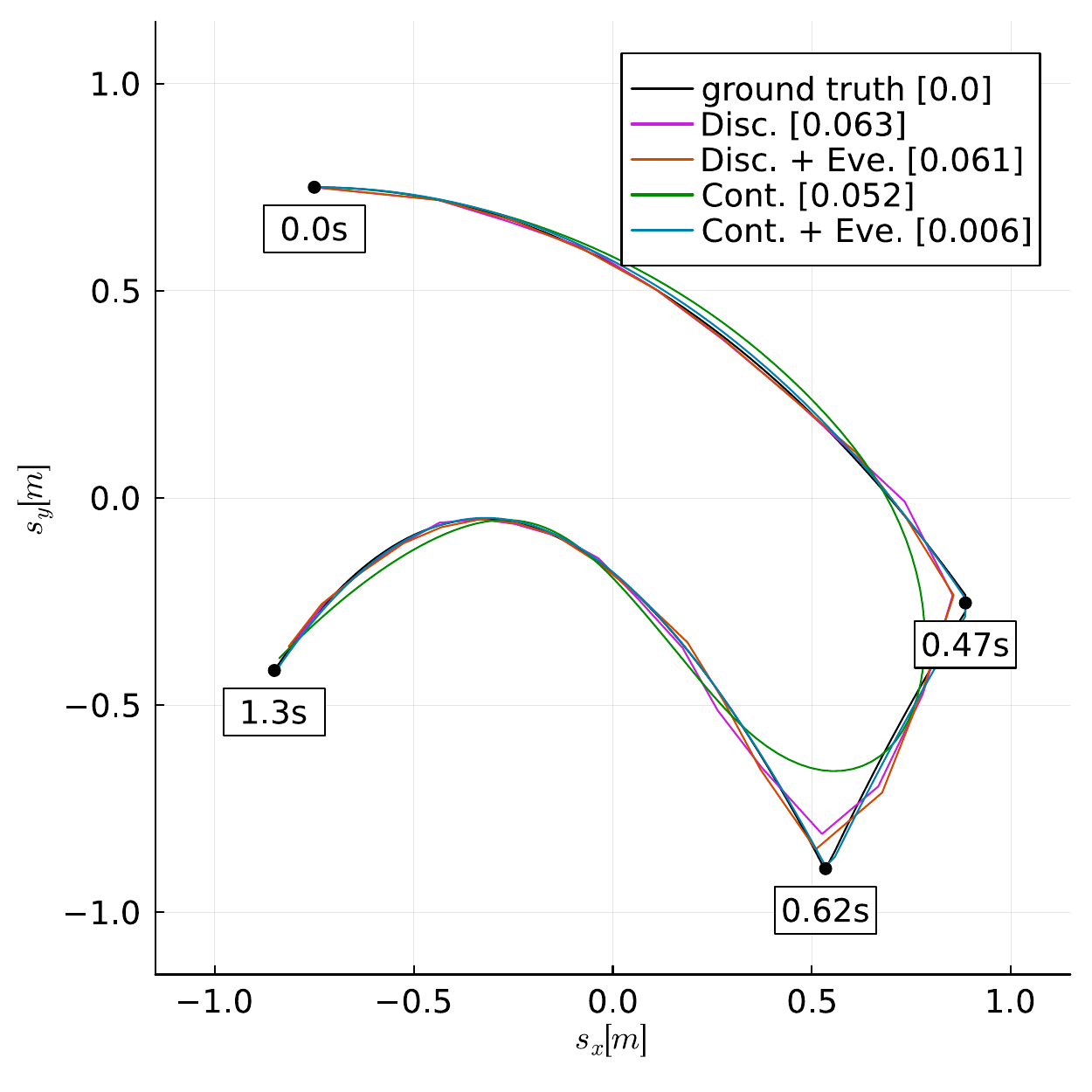}{Training results of the second experiment: The $s_x$-$s_y$-plane for the position of the ball on training data. Loss values (compared to ground truth) are in square brackets. Points in time for start, stop and collisions are marked with black dots and time labels to track the simulation trajectory over time.}{common}{0.5\textwidth}

As to expect, all four representatives of different model types can learn for the bouncing ball trajectory - within the scope of their individual possibilities. For example, the pure continuous neural \ac{ODE} is of course only able to approximate the non-differentiable parts of the trajectory.

\end{document}